%% file: main.tex
\documentclass[10pt,journal,compsoc]{IEEEtran}

\usepackage{booktabs} 

\usepackage{amsmath,epsfig}
\usepackage{color}
\usepackage{algorithm, algorithmicx}
\usepackage[noend]{algpseudocode}
\usepackage{epsfig, subfigure, multirow}
\usepackage{listings}
\usepackage{amsfonts}
\usepackage{bm}
\usepackage{pifont}
\usepackage{epstopdf}
\usepackage{hyperref}
\usepackage{multirow}
\usepackage{tabularx}
\usepackage[multiple]{footmisc}

\usepackage{lipsum}

\makeatletter
\algnewcommand{\LineComment}[1]{\Statex \hskip\ALG@thistlm \(\triangleright\) #1}
\makeatother

\newcommand{\sstitle}[1]{\vspace*{0.2em}\noindent{\bf #1\/.}}

\DeclareMathOperator*{\argmax}{arg\,max}

\newenvironment{myitem}{
\begin{itemize}
  \setlength{\parskip}{0pt}
  \setlength{\itemsep}{0pt}
  \setlength{\partopsep}{0pt}
  \setlength{\parskip}{0pt}
  \setlength{\topsep}{0pt}
  \setlength{\parsep}{0pt}}{\end{itemize}
}

\newenvironment{myenum}{
\begin{enumerate}
  \setlength{\parskip}{0pt}
  \setlength{\itemsep}{0pt}
  \setlength{\partopsep}{0pt}
  \setlength{\parskip}{0pt}
  \setlength{\topsep}{0pt}
  \setlength{\parsep}{0pt}}{\end{enumerate}
}

\newtheorem{mydef}{Definition}

\begin{document}

\title{Crowd-sensing Enhanced Parking Patrol \\ using Sharing Bikes' Trajectories}

\author{Tianfu He, Jie Bao, Yexin Li, Hui He and Yu Zheng.
    \IEEEcompsocitemizethanks{
      \IEEEcompsocthanksitem Tianfu He and Hui He are with Harbin Institute of Technology. Tianfu He is also with JD iCity, JD Technology and JD Intelligent Cities Research. E-mail: tianfudhe@foxmail.com and hehui@hit.edu.cn\protect
      \IEEEcompsocthanksitem Jie Bao, Yexin Li and Yu Zheng are with JD iCity, JD Technology and JD Intelligent Cities Research. E-mail: \{baojie3, liyexin\}@jd.com and msyuzheng@outlook.com\protect
      \IEEEcompsocthanksitem Hui He is the corresponding author.
    }
  }
    \markboth{IEEE Transactions on Knowledge and Data Engineering,~Vol.~TBD, No.~TBD, January~2021}
  {Shell \MakeLowercase{\textit{T. He et al.}}: Bare Demo of IEEEtran.cls for Computer Society Journals}
  
  \IEEEtitleabstractindextext{
  \begin{abstract}
    \input{abstract.tex}
  \end{abstract}
  
}

\maketitle

\input{intro}
\input{problem}
\input{preprocessing}
\input{detection}
\input{deploy}
\input{patrol}
\input{exp}

\input{related}
\input{conclusion}
\input{ack}

\bibliographystyle{IEEEtran}
\bibliography{sigproc} 

  \begin{IEEEbiography}[{\includegraphics[width=1in,height=1.25in,clip,keepaspectratio]{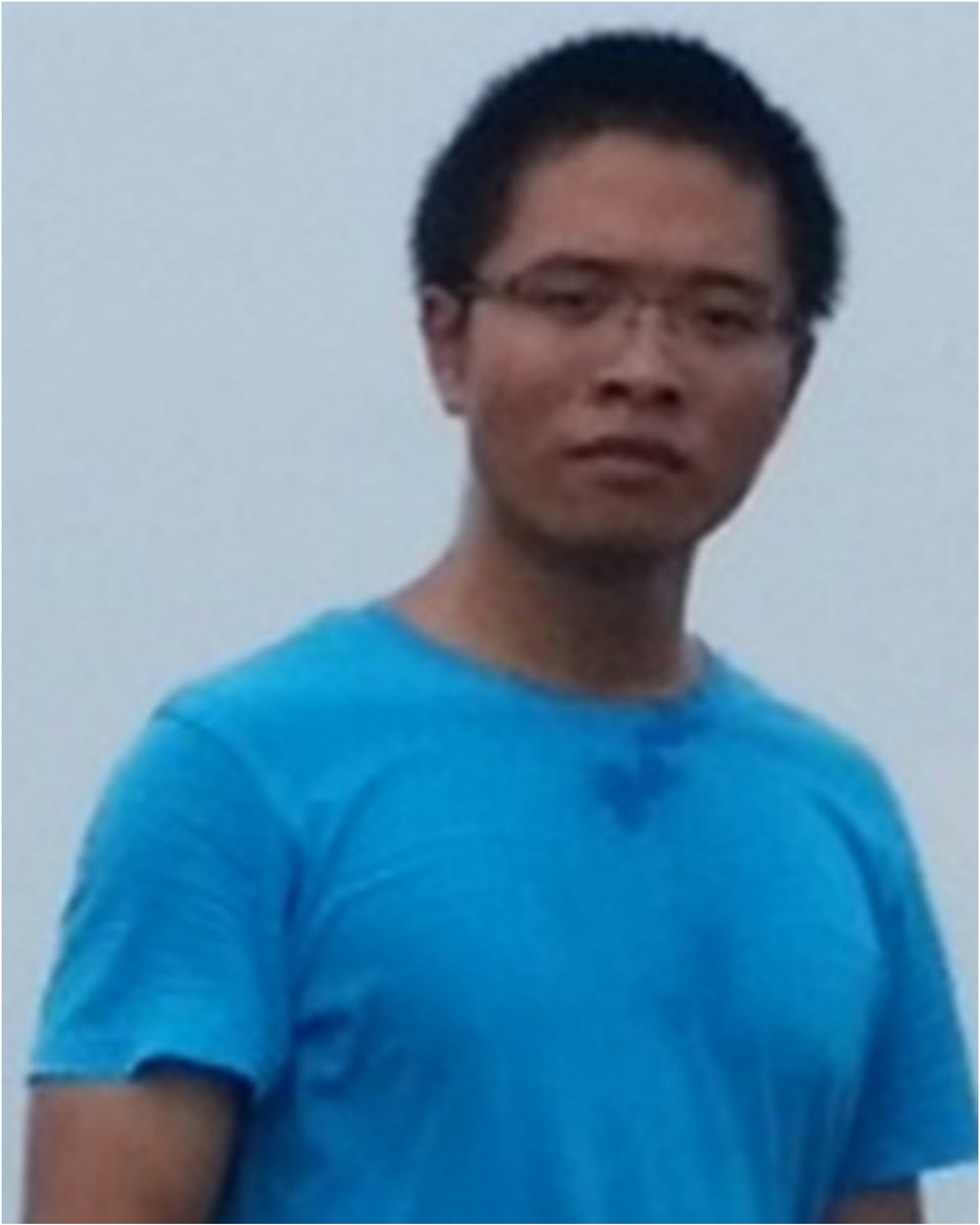}}]
  {Tianfu He} is currently a Ph.D. student in Computer Science Department, Harbin Institute of Technology. Before that, he received the B.E degree in Computer Science from Harbin Institute of Technology in 2016. His research interest involves urban computing, spatio-temporal data management, and data mining, especially trajectory data mining. He is passionate about new applied data science inventions, especially novel trajectory mining techniques for urban planning and management. 
  \end{IEEEbiography}

  \begin{IEEEbiography}[{\includegraphics[width=1in,height=1.25in,clip,keepaspectratio]{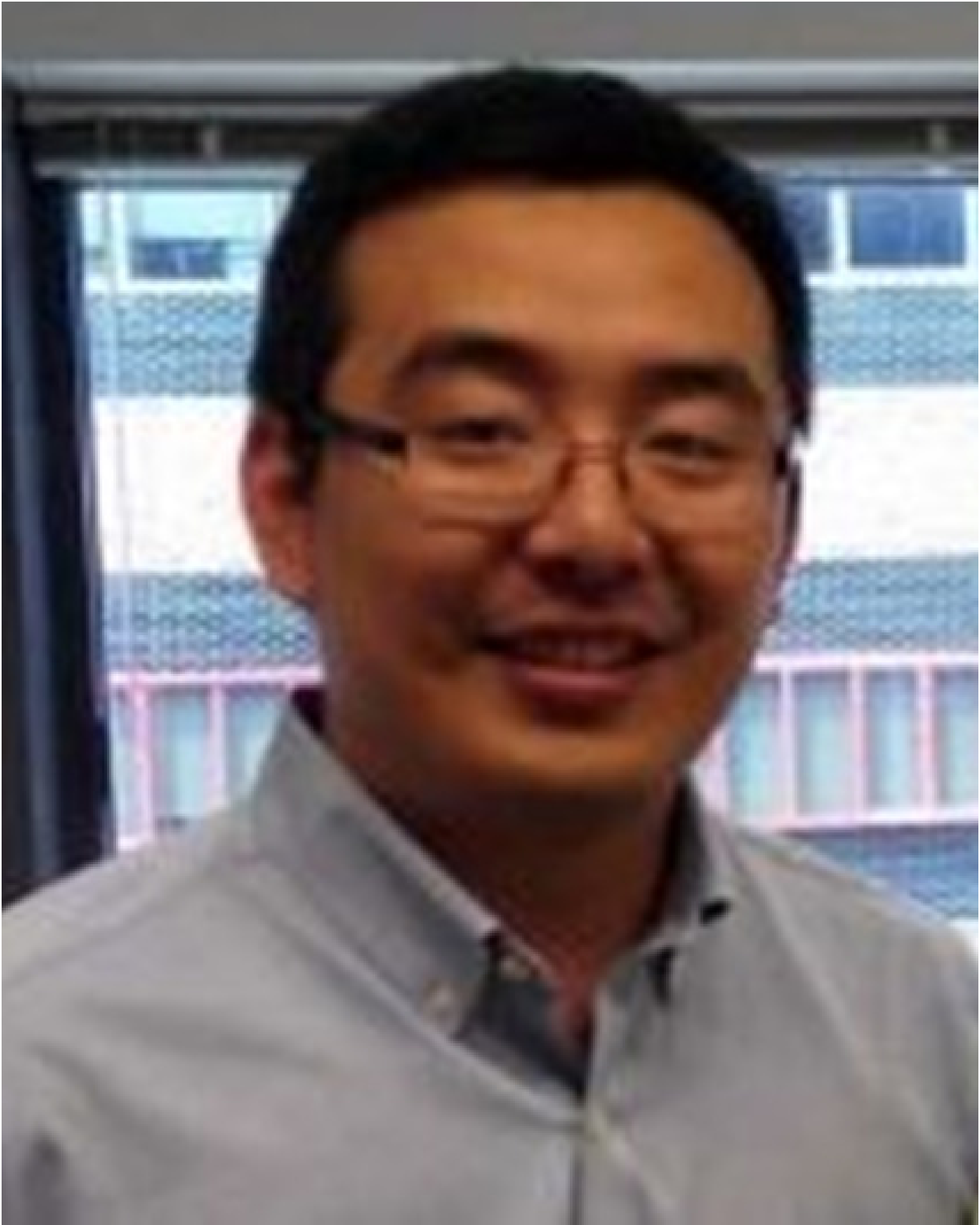}}]
  {Jie Bao} got his Ph.D degree in Computer Science from University of Minnesota at Twin Cities in 2014. He worked as a researcher in Urban Computing Group at MSR Asia from 2014 to 2017. He currently leads the Data Platform Division in JD Urban Computing Business Unit. His research interests include: Spatio-temporal Data Management/Mining, Urban Computing, and Location-based Services. 
  \end{IEEEbiography} 

  \begin{IEEEbiography}[{\includegraphics[width=1in,height=1.25in,clip,keepaspectratio]{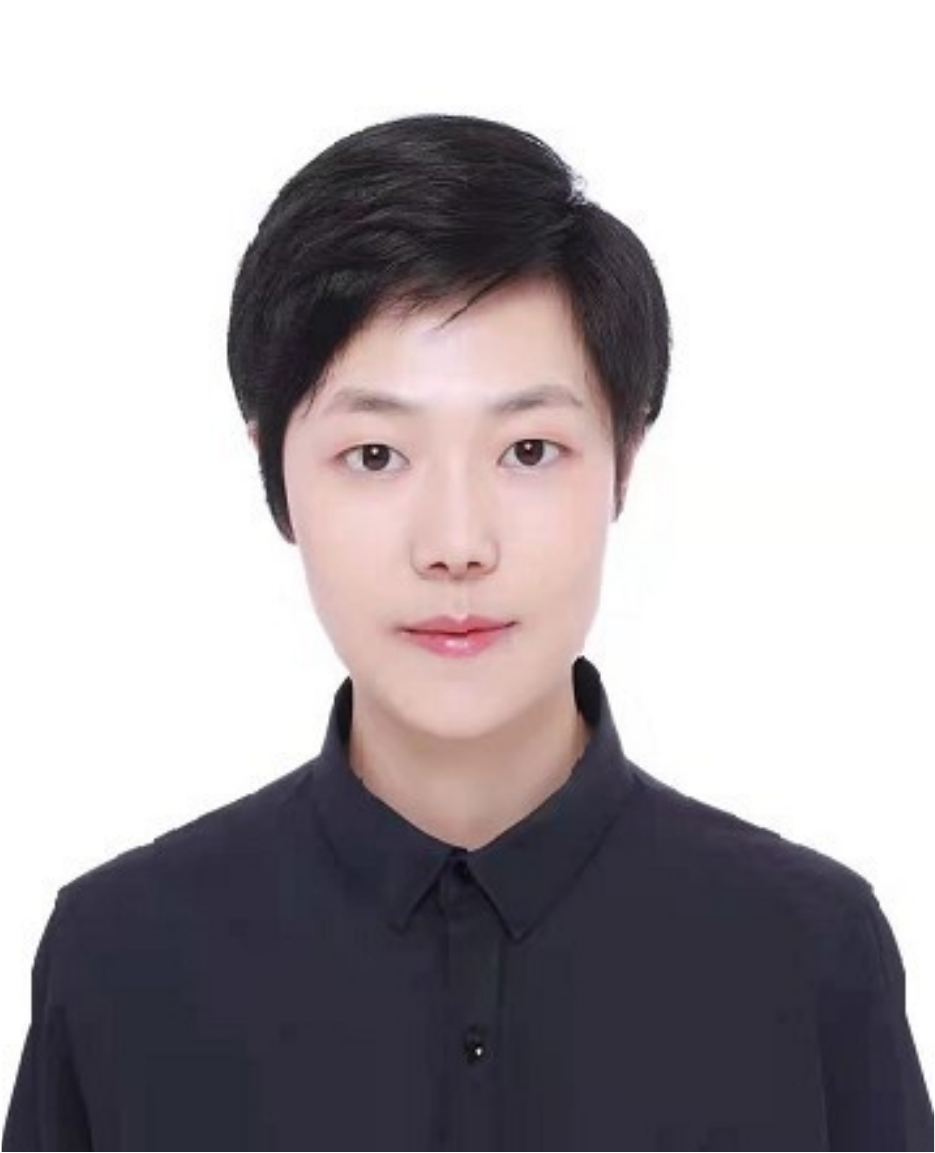}}]
  {Yexin Li} got her Ph.D degree in Computer Science and Engineering from the Hong Kong University of Science and Technology in 2021. She was an intern in Urban Computing group, MSR Asia. She is currently a researcher at JD Urban Computing Business Unit. Her research interests include deep reinforcement learning, spatio-temporal data mining, and urban computing.
  \end{IEEEbiography}

  \begin{IEEEbiography}[{\includegraphics[width=1in,height=1.25in,clip,keepaspectratio]{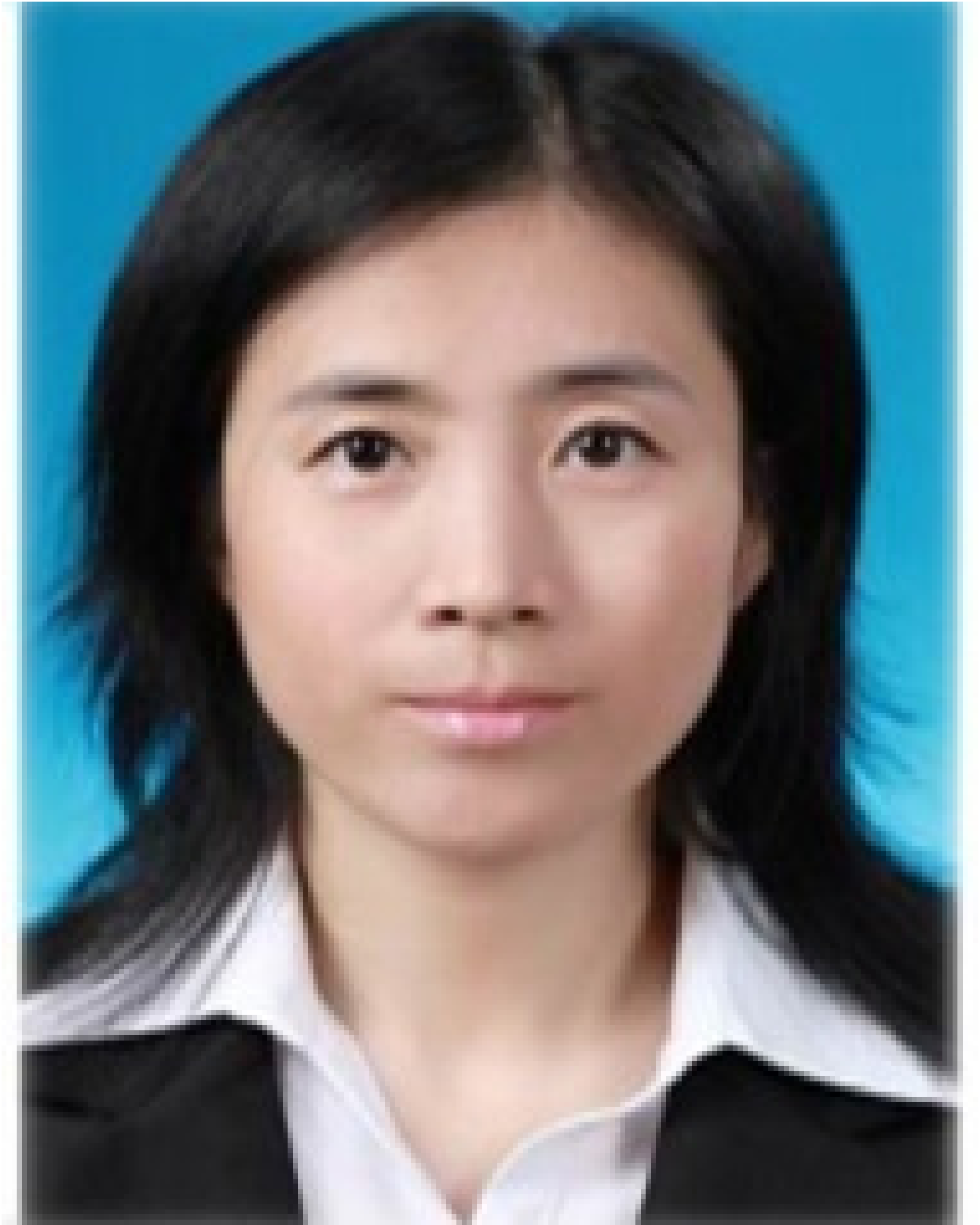}}]
  {Hui He} is currently Professor in School of Computer Science, Harbin Institute of Technology. She is a member of the IEEE, ACM and CCF. Her research interests includes information technology, big data processing and analysis and mobile network computing. She has accomplished many projects such as National High Technology Research and National Science Foundation Projects. 
  \end{IEEEbiography}

  \begin{IEEEbiography}[{\includegraphics[width=1in,height=1.25in,clip,keepaspectratio]{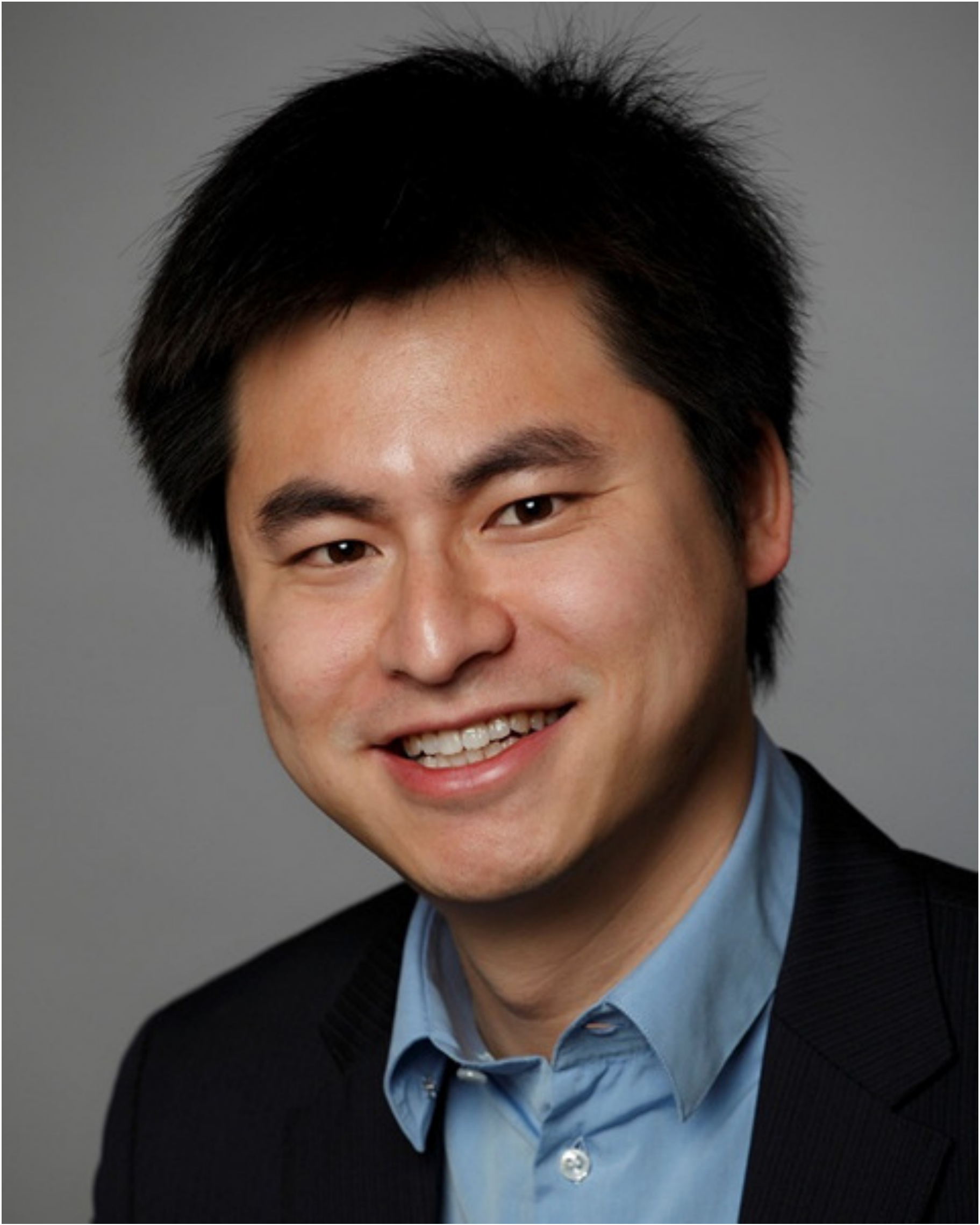}}]
  {Yu Zheng} is a Vice President and Chief Data Scientist at JD Finance Group, passionate about using big data and AI technology to tackle urban challenges. His research interests include big data analytics, spatio-temporal data mining, machine learning, and artificial intelligence. He also leads the JD Urban Computing Business Unit as the president and serves as the director of the JD Intelligent Cities Research. Before joining JD, he was a senior research manager at Microsoft Research. Zheng is also a Chair Professor at Shanghai Jiao Tong University, an Adjunct Professor at Hong Kong University of Science and Technology. He is Fellow of IEEE. 
  \end{IEEEbiography}

\end{document}

%% file: abstract.tex
Illegal vehicle parking is a common urban problem faced by major cities in the world, as it incurs traffic jams, which lead to air pollution and traffic accidents. 
The government highly relies on active human efforts to detect illegal parking events. However, such an approach is extremely ineffective to cover a large city since the police have to patrol over the entire city roads. 

The massive and high-quality sharing bike trajectories from Mobike offer us a unique opportunity to design a ubiquitous illegal parking detection approach, as most of the illegal parking events happen at curbsides and have significant impact on the bike users. The detection result can guide the patrol schedule, i.e. send the patrol policemen to the region with higher illegal parking risks, and further improve the patrol efficiency. Inspired by this idea, three main components are employed in the proposed framework:
1)~{\em trajectory pre-processing}, which filters outlier GPS points, performs map-matching, and builds trajectory indexes;
2)~{\em illegal parking detection}, which models the normal trajectories, extracts features from the evaluation trajectories, and utilizes a distribution test-based method to discover the illegal parking events;
and 3)~{\em patrol scheduling}, which leverages the detection result as reference context, and models the scheduling task as a multi-agent reinforcement learning problem to guide the patrol police.
Finally, extensive experiments are presented to validate the effectiveness of illegal parking detection, as well as the improvement of patrol efficiency.

%% file: intro.tex
\section{Introduction}

\begin{figure}[t]
\begin{center}
\includegraphics[width=3.4in]{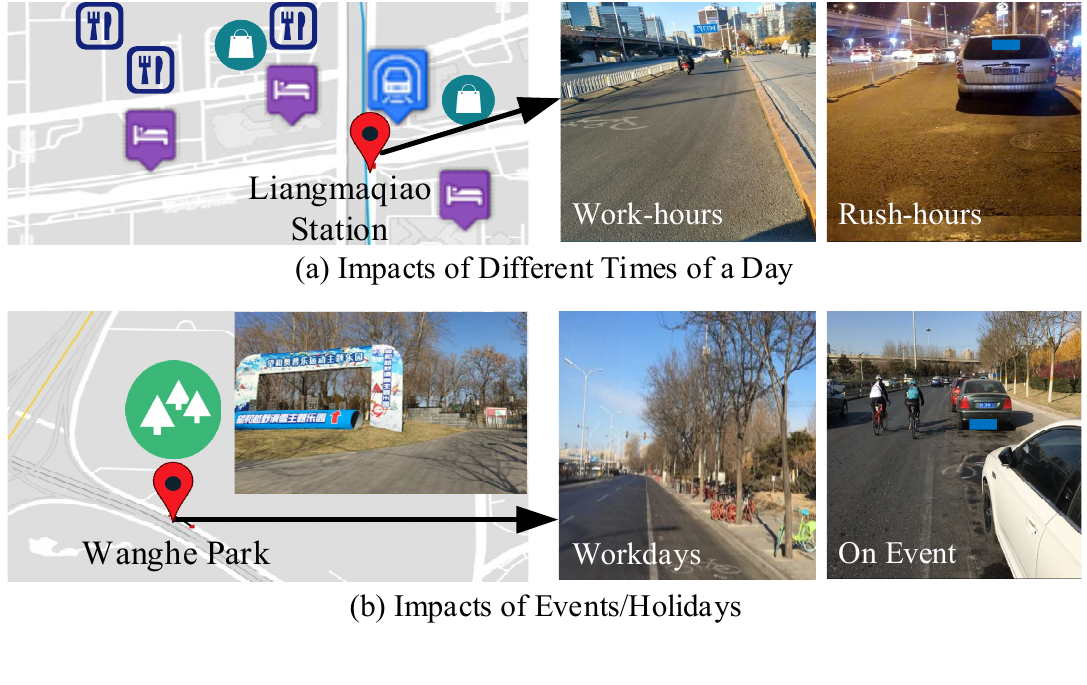}
\vspace{-10pt}
\caption{Dynamics of Illegal Parking. Parking violations in (a) during rush-hours are caused by didi/uber drivers or nearby restaurant customers; and (b) shows Wanghe Park's snow festival, which attracts many kids and their families.}
\label{fig:motivation}
\end{center}
\end{figure}

Illegal vehicle parking is a common problem in large cities all over the world. The illegal parking events decrease the transportation efficiency in a city, and incurs traffic jams~\cite{MGC12}, which lead to air pollution~\cite{KA13} and potential accidents (as illustrated in Figure~\ref{fig:motivation}b, bike users have to ride on the vehicle lanes). Fine ticketing is the major approach for urban governors to deal with illegal parking. The policemen in team groups are responsible for patrolling the roads over the city and pasting fine tickets on illegal vehicles. However, such an approach requires heavy human efforts to ensure wide patrol coverage, since the illegal parking events are caught only when they are in sight. Besides, the patrol efficiency can even be worse due to the high dynamics of the illegal parking situation(Figure~\ref{fig:motivation}), especially the non-periodical dynamics caused by temporary activities in Figure~\ref{fig:motivation}b. Nevertheless, untargeted patrolling can hardly achieve a satisfactory effect, yet is very tedious for the police. With the advances of video object identification technologies, some research works~\cite{LRR+09,TFL+11} identify the illegal parking events based on surveillance cameras. However, it is costly to achieve a high camera coverage level in large cities. For example, in Beijing there are only 445 cameras for illegal parking detection~\footnote{\url{https://www.icauto.com.cn/weizhang/wzd/110100/4.html}}. To this end, in this paper, we propose a novel and ubiquitous approach to effectively detect illegal vehicle parking events by mining the trajectories of sharing bikes, based on which a patrol scheduling framework is built to guide the police.

\begin{figure}[t]
  \begin{center}
  \includegraphics[width=3.5in]{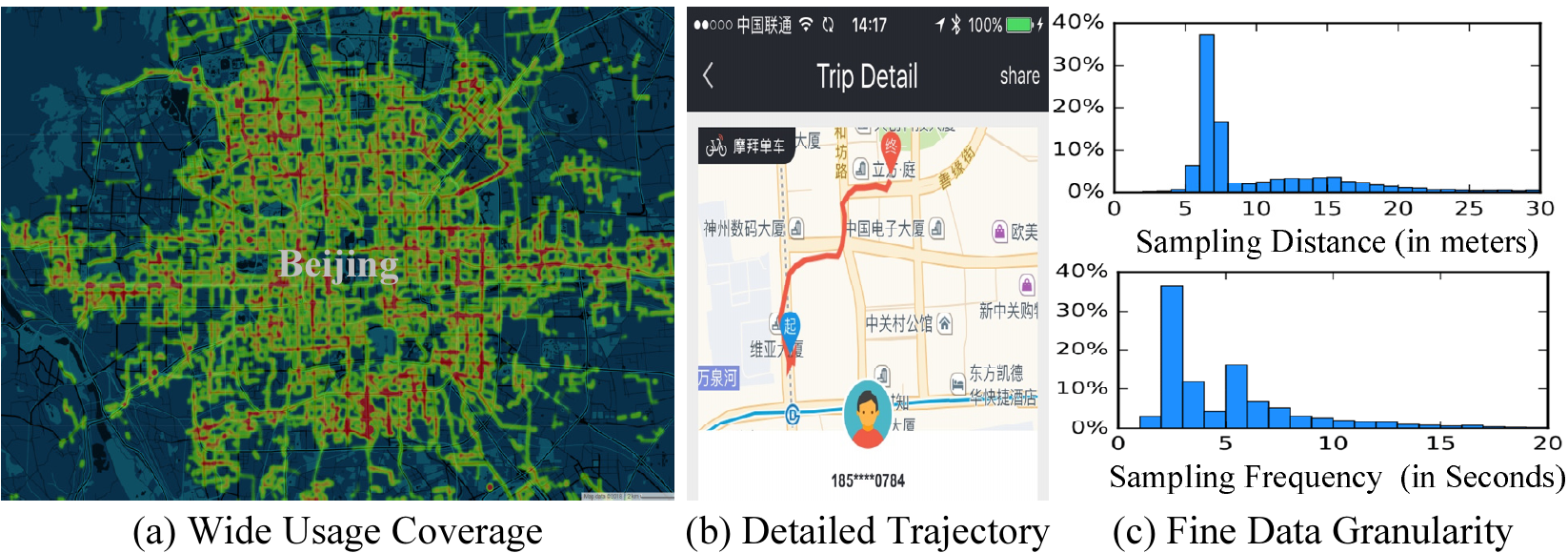}
    \caption{Opportunity in Mobike Trajectories.}
  \label{fig:opportunity}
  \end{center}
  \vspace{-15pt}
\end{figure}

\sstitle{Opportunity} The intuition behind the detection technique is that, based on our observation, illegal vehicle parking events usually take place at curbsides, which block the path of bike users and significantly affect their trajectories. Therefore, by aggregating massive bike trajectories on the same road, we are able to identify the illegal parking events via examining the distinct patterns of their trajectories. Fortunately, the bike trajectory data provided by Mobike~\footnote{https://en.wikipedia.org/wiki/Mobike}\footnote{Mobike is currently acquired by Meituan-Dianping} (a station-less sharing bike service provider in China), offers us a unique opportunity to tackle the detection problem with two distinctive advantages:

\begin{myitem}
\item \textbf{Wide Usage Coverage.} Mobike is a very popular bike sharing service, which is frequently used as a daily commute mode for many people nowadays. According to the recent report~\cite{MobikeScale}, it got more than 200 million registered users and 30 million daily trips. Moreover, Mobike trajectories cover widely across a city, e.g., Figure~\ref{fig:opportunity}a visualizes the Mobike usages in the city of Beijing. With the most roads in the urban area densely covered, it is possible for us to detect illegal parking events in large cities without any active efforts. 

\item \textbf{High Data Quality.} First of all, Mobike records detailed GPS trajectories for each trip, as demonstrated in Figure~\ref{fig:opportunity}b. Moreover, the granularity of each trajectory is very high, as shown in Figure~\ref{fig:opportunity}c: 1)~more than 60\% of the distances between two GPS points are less than 6 meters, and 2)~over 70\% of the time interval between two GPS points are less than 6 seconds. Therefore, it is possible for us to identify the subtle traveling behavior changes, caused by vehicle illegal parking events.
\end{myitem}

\sstitle{Intuition Validation} With the access to the large-scale and high-quality Mobike trajectories, we first conduct a set of experiments to validate the feasibility of our intuition, i.e., whether it is possible to identify illegal parking events based on sharing bike trajectories. We ride a Mobike on a local road (as shown Figure~\ref{fig:intuition}a multiple times, where the area marked in the white lines is the simulated illegal parking location), with two settings: 1)~with simulated illegal parking vehicles, and 2)~without simulated illegal parking vehicles, each for ten times, i.e., conceptually demonstrated in Figure~\ref{fig:intuition}b.   

Figure~\ref{fig:intuition}c visualizes the experimental trajectories extracted from Mobike, where the red lines are the trajectories with illegal parking simulation, and the blue lines are the normal trajectories. It is clear that, especially around the simulated illegal parking location (marked with the orange circle), comparing to the normal trajectories, the affected trajectories are more twisted and leaning toward the opposite side of the curbside. As a result, this set of experiments confirms our intuition.

\sstitle{Patrol Scheduling} The detection capability of sharing bike trajectories not only helps the patrol of individual police, but also contributes to the city-wide scheduling of the patrol teams: by aggregating the road-granular detection results, we can get the approximate distribution over the city, and put in more patrol police to regions with higher risks. According to the studies in~\cite{VAT+17}, due to the lack of social funding, it is common that cities only have limited patrol efforts. This urges the government to promote a more efficient patrol scheduling framework to improve the utilization of patrol resources. In this paper, we partition the city regions into grids, and at each time step, the designed scheduling algorithm determines where each policeman should go and patrol. The detailed formulation of the patrol scheduling framework will be elaborated later in this paper.

\sstitle{Challenges and Ideas} However, to realize the ideas all above, there are still many challenges in three aspects: 1)~{\em complex data nature}. There are data errors caused either by the GPS module or human errors (e.g., forget to lock and return bike). Besides, map-matching is a crucial step to match trajectories to correct road segments. It is more difficult working with bike trajectories, as bikes can be ridden more freely beyond the roads; 2)~{\em illegal parking detection}, developing an effective detection model is not trivial, as it is hard to collect massive labeled data. The efficiency is another important issue. With the massive trajectories in a large city, the system response time needs to be minimized for users (i.e., city managers) to identify all illegal parking events in a city; 3)~{\em patrol scheduling optimization}. As is discussed in Figure~\ref{fig:motivation}, the complex dynamics of illegal parking makes it difficult for the scheduling system to optimize the long-term patrol effectiveness. In addition, as there are multiple patrol teams, the system should also consider the collaboration of them.

In this paper, we design, implement an parking patrol scheduling framework based on data mining results from the massive sharing bikes' trajectories. The system consists of three main modules: 1)~{\em pre-processing}, which filters outlier GPS points, performs map-matching, and builds indexes; 2)~{\em illegal parking detection}, which studies a baseline to model the normal trajectories for each road, extracts the features of evaluation trajectories and infers the possibility of the presence of illegal parking events. To improve the system response time and efficiency, the trajectory data is stored on a distributed storage platform, i.e., MongoDB, and the illegal parking detection system is deployed on Apache Storm; and 3)~{\em reinforcement learning-based scheduling}. By formulating the patrolling task as a Markov Decision Process, we propose to solve the long-term scheduling problem by collaborative multi-agent reinforcement learning. The main contributions of the paper are summarized as follows:

\begin{figure}[t]
  \begin{center}
  \includegraphics[width=3.5in]{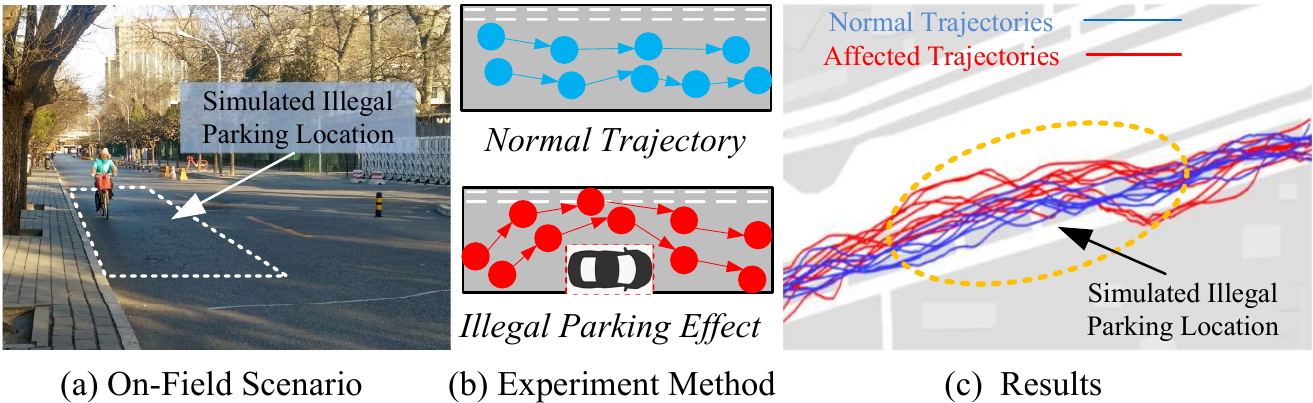}
    \caption{Intuition Validation Experiments.}
  \label{fig:intuition}
  \end{center}
  \vspace{-15pt}
\end{figure}
\vspace{5pt}

{$\bullet$} We provide the first attempt on detecting illegal parking events ubiquitously by mining massive bike trajectories.

{$\bullet$} We design and implement a comprehensive {\em preprocessing} module to clean bike trajectories, map them to corresponding road segments and build a set of indexes. We also propose a novel distribution test-based approach to detect the illegal parking events on a road segment, and deploy the algorithm on distributed system to improve the efficiency.

{$\bullet$} We collected over 400 illegal parking labels manually to tune the most effective threshold in the detection model.

{$\bullet$} We evaluate the proposed model extensively over six months' Mobike trajectory data from the City of Beijing. Moreover, on-site case studies are conducted to validate the effectiveness of our illegal parking detection. 

{$\bullet$} To the best of our knowledge, this is the first parking patrol framework that is driven by the passive crowd-sensing detection results. It provides us with an opportunity to formulate the patrolling problem as a context-aware multi-agent reinforcement learning problem. Both the idea and the model are validated effective in our comprehensive experiments.
\vspace{5pt}

The rest of the paper is organized as follows: Section~\ref{sec:problem} describes the problem and the system overview. Section~\ref{sec:preprocessing} presents the pre-processing module. Illegal parking detection is discussed in Section~\ref{sec:detection}. Section~\ref{sec:deploy} presents the system deployment to improve detection efficiency. The formulation of patrol scheduling and our solution are given in Section~\ref{sec:patrol}, where the patrol simulator is also presented, which is used to train our model and evaluate the effectiveness of patrol scheduling methods. Experiments and case studies are given in Section~\ref{sec:exp}. Related works are summarized in Section~\ref{sec:related}. Finally, Section~\ref{sec:conclusion} concludes the paper.

%% file: problem.tex
\section{Overview}\label{sec:problem}
\subsection{Preliminaries}

\begin{mydef}
\label{def:gps_traj}
\textbf{(Trajectory)} A trajectory $\tau$ can be defined as a time-ordered sequence $\tau = \{p_1 \to p_2 \to ... \to p_n\}$, where $p_i = (lat_i, lng_i, t_i), 1 \le i \le n$, is a GPS record with latitude $lat_i$, longitude $lng_i$, and timestamp $t_i$.
\end{mydef}
The Mobike trajectory starts when the user scans the QR code to unlock a sharing bike, and ends when the user locks the bike. The intermediate GPS points are sampled at a constant rate.

\begin{mydef}
\label{def:road_network}
\textbf{(Road Network)} A road network $RN$ is a directed graph $G$ = ($V$, $E$), where $V = \{v_1, v_2, ..., v_m\}$ is a set of intersections, and $E = \{e_1, e_2, ..., e_n\}$ is a set of road segments (edges).
For each $e_i \in E$, it associates with three properties: 1)~{\em level}, which indicates the type of road;
2)~{\em shape}, which is a sequence of location points, from a \texttt{FromNode} to a \texttt{ToNode}, describing the shape of the segment; and 2)~{\em dir}, which indicates its directional information (bidirectional or uni-directional).
\end{mydef} 

\begin{figure}[t]
  \begin{center}
  \includegraphics[width=3.4in]{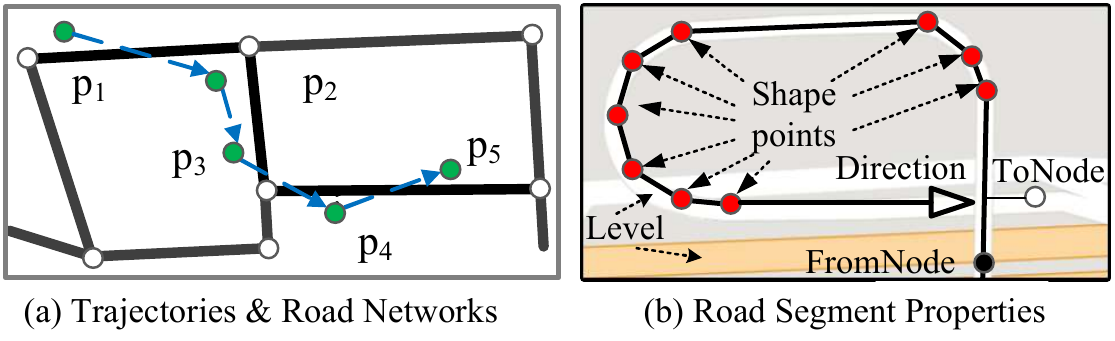}
    \caption{Examples of Preliminary Concepts.}
  \label{fig:prelimexample}
  \end{center}
  \vspace{-15pt}
\end{figure}

Figure~\ref{fig:prelimexample}a gives an example of the concepts. The green dots are GPS points, and the blue arrows indicate their sequence. On the other hand, the white dots are the nodes and lines are the edges of the road network. Figure~\ref{fig:prelimexample}b illustrates the detailed properties of a road segment, where the red dots are the location points describing the {\em shape}, the white dot and the black dot are the  \texttt{FromNode} and \texttt{ToNode} respectively, the arrow indicates its {\em dir} property (unidirectional in this case), and the colors represent different levels of roads, e.g., the yellow color represents highways, and the white color means supplementary roads.

\begin{mydef}
\textbf{(Illegal Parking Event)} An illegal parking event in this paper refers to obstacles at a road segment $e_i$ that affects the normal behavior of bike trajectories ($Tr_{e_i}$) on it, during a temporal range $t_i$ \& $t_{i+1}$ (e.g., 8:00 AM to 9:00 AM).
\end{mydef}

\begin{mydef}
\textbf{(Illegal Parking Detection)} Given a set of trajectories $Tr$, a road network $G=(V,E)$, and temporal period $t_i$ \& $t_{i+1}$, for every road segment (or edge $e_i \in E$), we want to infer the possibility of the presence of the illegal parking events, based on the sharing bikes' trajectories generated on each road segment $e_i$ from $t_i$ \& $t_{i+1}$. 
\end{mydef}

\begin{figure}[b]
  \begin{center}  
  \vspace{-10pt}
  \includegraphics[width=1.05\linewidth]{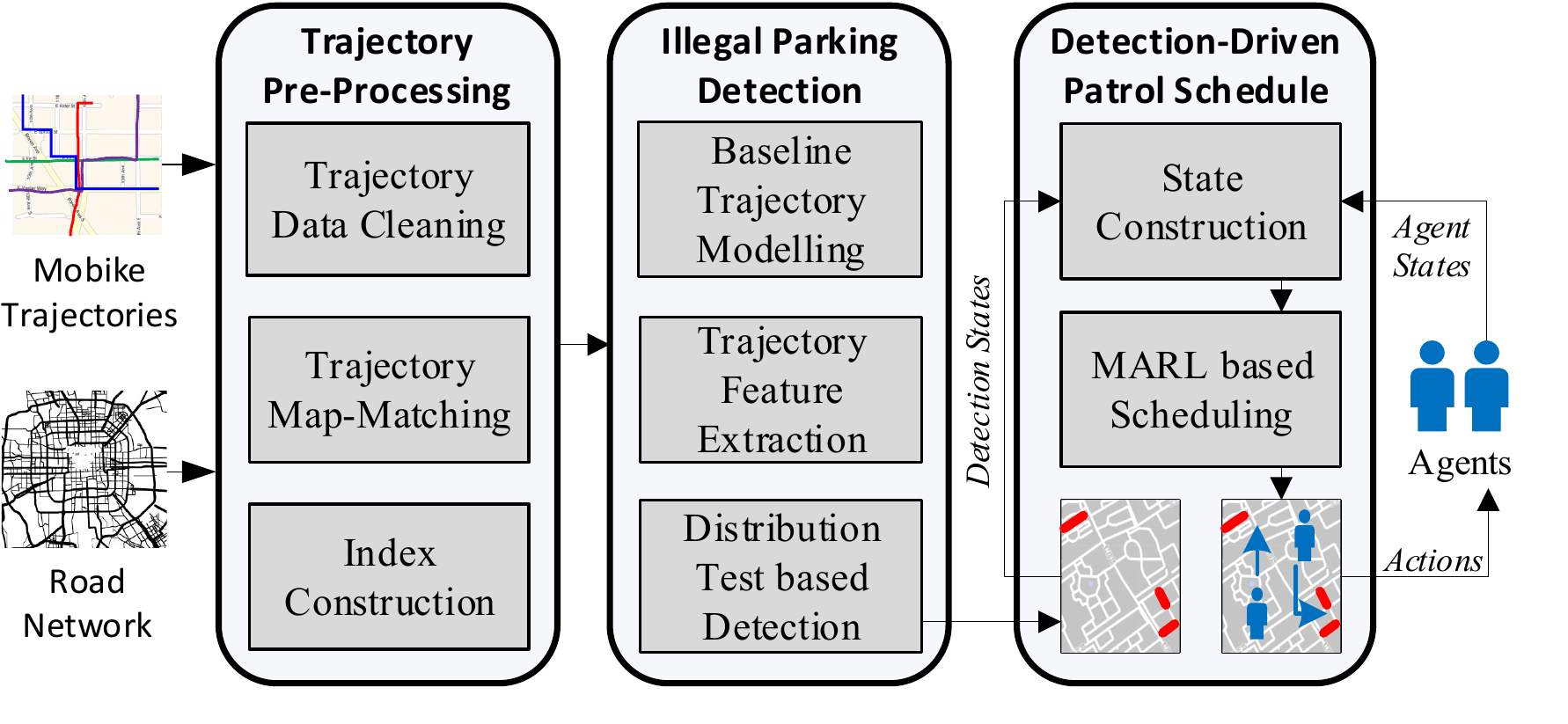}
    \caption{System Overview.}
  \label{fig:overview}
  \end{center}
\end{figure}

\subsection{System Overview}

Figure~\ref{fig:overview} gives a system overview with three main components:

\sstitle{Pre-Processing} This component takes bike trajectories and road networks and performs three main tasks: 1)~{\em Trajectory Data Cleaning}, which removes the outlier GPS points; 2)~{\em Trajectory Map-Matching}, which projects GPS points onto the corresponding road segments; and 3)~{\em Index Construction}, which builds indexes to speed up the trajectory retrieval process based on road segment IDs and temporal ranges (detailed in Section~\ref{sec:preprocessing}).

\sstitle{Illegal Parking Detection} This component calculates a score for each road segment, indicating the probability of the presence of illegal parking events, by evaluating the processed trajectories in a temporal period. Three main tasks are performed: 1)~{\em baseline trajectory modeling}, which builds a model for each road segment to describe the normal trajectories; 2)~{\em trajectory feature extraction}, which extracts the features from the evaluation trajectories; and 3)~{\em distribution test-based detection}, which detects illegal parking events using distribution tests (detailed in Section~\ref{sec:detection}).

\sstitle{Detection-Driven Patrol Scheduling} This component is driven by the real-time detection results and it decides where each patrol police should go and serve. By partitioning the city into grids, the patrol system is formulated as Markov Decision Process, and solved by collaborative Multi-Agent Reinforcement Learning (MARL): the module learns the value function by Deep Q-Network by optimizing the long-term number of patrol catches, and the next actions for each agent are generated according to the learned value function one-by-one. The details are described in Section~\ref{sec:patrol}.

%% file: preprocessing.tex
\section{Trajectory Pre-Processing}\label{sec:preprocessing}

As the data quality of bike trajectories used in our system determines the accuracy of the illegal parking detection, a set of pre-processing tasks are necessary, before the massive trajectories from Mobike users can be used: 1)~{\em Trajectory Data Cleaning}, which removes the GPS outliers in a trajectory based on the speed and sampling rates; 2)~{\em Trajectory Map-Matching}, which segments the GPS points in the trajectories and maps them onto the corresponding road segments; and 3)~{\em Index Construction}, which builds the indexes to speed up the trajectory data retrieval process.

\subsection{Trajectory Data Cleaning}

\begin{figure}[t]
  \begin{center}
  \includegraphics[width=3.4in]{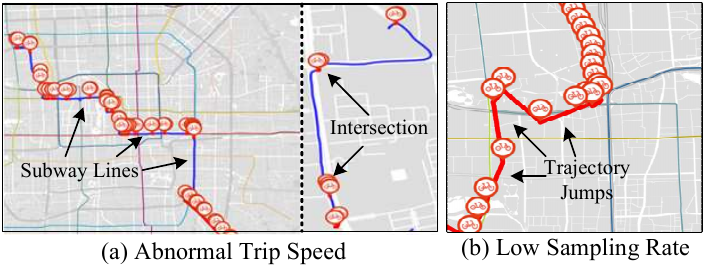}
    \caption{Typical Examples of Data Errors: the markers in (a) are the sampled trajectory points with abnormally high/low speed; and (b) gives a trajectory with extremely low sampling rate.}
  \label{fig:datacleaning}
  \end{center}
  \vspace{-15pt}
\end{figure}

This module cleans the raw trajectories from Mobike. Essentially as a type of crowd sensing data, Mobike trajectories are generated by the GPS modules from mobile phones. As a result, a noticeable portion of trajectories have different data errors, which significantly affect the accuracy of illegal parking detection:

\begin{myenum}

\item \textbf{Abnormal Speeds.} As most users ride bikes at normal speeds (e.g., 5 kmph to 20 kmph), there are two types of abnormal speeds: 1)~{\em abnormal high speed}, which is caused by GPS errors, or unusual usage (e.g., demonstrated as the left portion of Figure~\ref{fig:datacleaning}a, where a user got on a subway without locking the bike); and 2)~{\em abnormal low speed}, as is demonstrated at the right part of Figure~\ref{fig:datacleaning}a, which is usually caused by the traffic lights at intersections (crossroads), instead of the parking violations at curbside.

\item \textbf{Low Sampling Rates.} In some cases, due to the errors of GPS modules in users' mobile phones, some of the GPS points may be missing. Figure~\ref{fig:datacleaning}b shows an example of a bike trajectory travelling with a normal speed, but with several jumps (marked in red lines). 
\end{myenum}

Both of the above data quality issues introduce problems in the detection model. For example, the trajectory segments with abnormal low speed can be affected by many factors other than illegal parkings, e.g., pedestrians. Moreover, the trajectory segments with low sampling rates introduce challenges in map-matching and provide no information reflecting the conditions of the roads. As a result, a heuristic based approach~\cite{Z15} is used here to clean the trajectory data: If any portion of the consecutive GPS points fails to meet the speed or sampling rate thresholds, these disqualified segments are removed from the original trajectory. In the end, one trajectory may be segmented into several short qualified sub-trajectories to preserve more information.

\subsection{Trajectory Map-Matching}\label{sec:preprocessing:mm}

In this module, we map the GPS points onto the corresponding segments in road networks, which is crucial for the illegal parking detection. Traditional map-matching algorithms, e.g.,~\cite{YZZ+10}, cannot be used directly, because, comparing to vehicle trajectories, bike trajectories have several unique properties: 1)~they have much lower travelling speeds, 2)~they travel at both directions even at a uni-directional road, 3)~they can go to the area without road networks; and 4)~they have more short trips. To adapt with these properties, the map-matching module is designed with three steps:

\begin{figure}[t]
  \begin{center}
  \includegraphics[width=3.4in, height=1.0in]{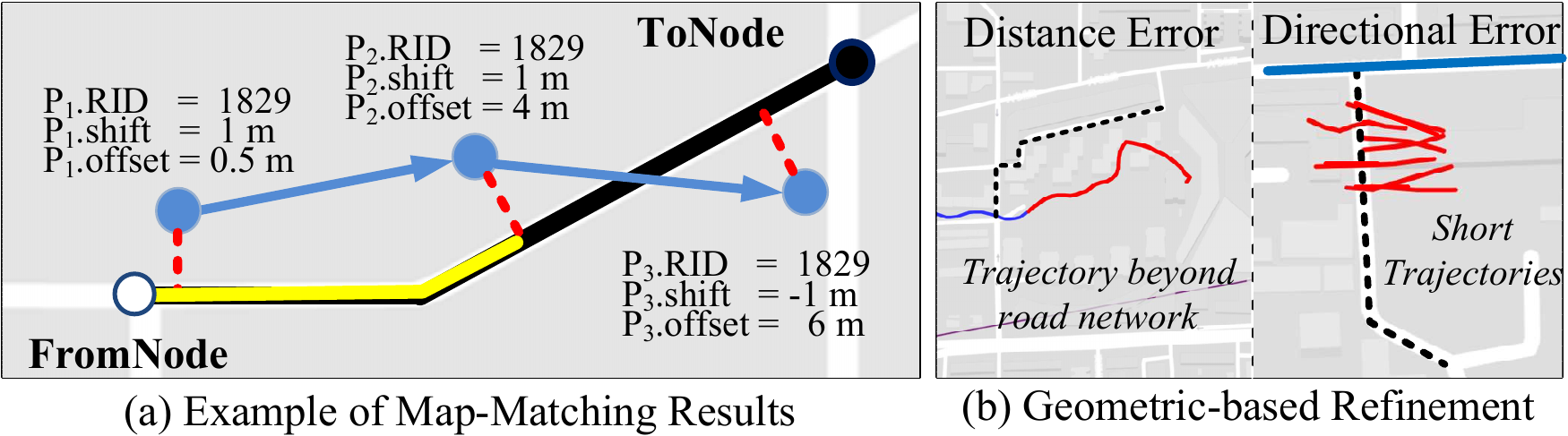}
    \caption{Trajectory Map-Matching. (a): the result of map-matching, with each trajectory point associated with three properties, i.e. RID, shift and offset; and (b): two typical map-matching errors, which are caused by the unique properties of bike trajectories.}
  \label{fig:mapmatching}
  \end{center}
    \vspace{-15pt}
\end{figure}

\sstitle{Step 1. Adaptive Map-Matching} This step employs an interactive-voting based map matching algorithm~\cite{YZZ+10} with three modifications: 1)~high level roads, which can be only used by vehicles (e.g., highways), are removed; 2)~the direction information on road segments is omitted, and all road segments are set as bidirectional; and 3)~the speed constraint of each road segment is not used to adapt the slower speed in bike trajectories.  

After the map-matching process, as shown in Figure~\ref{fig:mapmatching}a, each GPS point is associated with three new properties: 1)~{\em RID}, which is the map-matched road segment ID; 2)~{\em shift}, which is the shortest distance to the road segment, illustrated as the red dotted lines. We define the positive shift for the GPS points at the left side of a road segment (direction is from \texttt{FromNode} to \texttt{ToNode}, as $P_1 \& P_2$), and negative shift at the right side of the road (as $P_3$); and 3)~{\em offset}, which is the length between the \texttt{FromNode} and the projection of the GPS point, as the yellow segment is the offset for $P_2$.

\sstitle{Step 2. Geometric-based Refinement} This step removes the problematic map-matching results, using geometric filters. Figure~\ref{fig:mapmatching}b shows two types of problematic map-matching results: 1)~{\emph distance error}, which is caused by the incompleteness of the road network data. As demonstrated in the left portion of Figure~\ref{fig:mapmatching}b, the algorithm maps the trajectory inside a residential area onto the black dotted road segments around it. It is because the road network we used is not detailed enough to reflect these small roads; and 2)~{\em directional error}, which is caused by the short trips in the data set, generated by the users or as the result of the trajectory data cleaning process. The right portion of Figure~\ref{fig:mapmatching}b shows many short trips (marked in red), which are mapped onto the black dotted road segment, while it makes more sense to map them onto the blue segment, as they head similar directions.

\begin{figure}[t]
  \begin{center}
  \includegraphics[width=3.4in, height=1.0in]{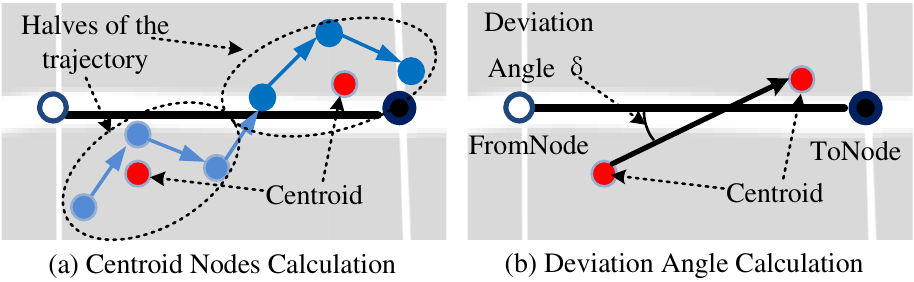}
    \caption{Map-Matching Refinement.}
  \label{fig:refinement}
  \end{center}
  \vspace{-15pt}  
\end{figure}

We use a geometric-based refinement to remove these errors, with the consideration of random shifts incurred by the GPS sensors. First of all, the distance errors are removed, if the average shift a sub-trajectory is greater than a threshold (e.g., 20 meters in our implementation). To remove the directional errors, a deviation angle is calculated between the directions of the overall trajectory and the road, as demonstrated in Figure~\ref{fig:refinement}. The overall trajectory direction is calculated by connecting the centroid points (i.e., the red dots) between the first and second portion of the sub-trajectory (i.e., circled by the dotted ovals). The trajectory is removed, if the deviation angle $\delta$ is greater than $\frac{\pi}{3}$.

\sstitle{Step 3. Reverse Trajectory Removal} This step removes the trajectories travelling at the reverse direction of the uni-directional roads. As all the roads are considered as bidirectional in the map-matching step, for a uni-directional road, there are a small number of reverse travelling trajectories by the users disobeying the traffic rules. Although the number of reverse travelling trajectories is limited, they usually have higher shift values, as they are much likely to encounter obstacles other than illegal parking events, e.g., bikes travelling at the normal direction. Therefore, the reverse trajectories affect the accuracy in our illegal parking detection model (experiments are provided).

\begin{figure}[h]
  \begin{center}
  \includegraphics[width=3.4in]{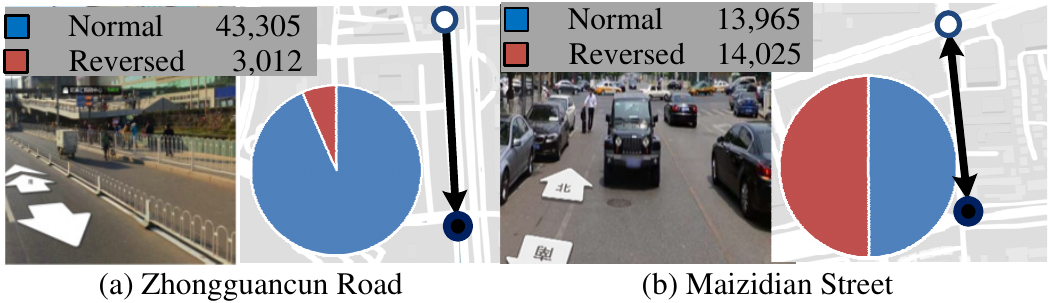}
    \caption{Results of Reverse Trajectory Identification.}
  \label{fig:direction}
  \end{center} 
  \vspace{-15pt} 
\end{figure}

To identify the reverse travelling behaviours, we calculate the overall direction of a trajectory by comparing the average offset between the two halves of the trajectories.  On a uni-direction road, if the average offset of the first half is less than it is in the second half, the trajectory travels reversely on the road segment. Figure~\ref{fig:direction} shows the distribution of normal and reversed trajectories we identified on a uni-direction road, i.e., Zhongguancun Road and a bi-directional road, i.e., Maizidian Street. The reverse trajectory identification result is consistent with our intuition, where much less numbers of reverse trajectories (less than 10\%) appear in a uni-direction road. On the other hand, the distribution is more balanced (i.e., 50\% for each direction) on the bidirectional road. Finally, all the reverse trajectories are removed.

\subsection{Index Construction} In this module, the system builds an {\em inverted index} based on each directional road segment (i.e., two entries are built for a bidirectional road segment), where each entry associated with the trajectories are mapped to it. Moreover, a temporal index is built based on the time stamp when the trajectory enters the road segment, as most of the latter trajectory retrieval tasks are based on the road segment ID and a temporal range. In our implementation, all trajectories and indexes are maintained in a MongoDB.

%% file: detection.tex
\section{Illegal Parking Detection}\label{sec:detection}


\subsection{Overview}

\sstitle{Challenges} With the massive, high quality, and pre-processed bike trajectories, detecting illegal parking events on a road segment is still a very hard problem: 

\vspace{5pt}
\begin{myenum}
\item~{\em No labeled data.} We do not have large scale labels for illegal parking events, which makes it hard to apply the conventional classification models directly. 

\item~{\em Complex illegal parking events.} The scenarios of illegal parking events are various, even on the same road, as they appear with different numbers and in different positions.

\item~{\em Variant individual behaviors.} Users have different riding preferences and behaviors, which makes it unstable to infer illegal parking events with an individual trajectory.

\item~{\em GPS inaccuracy.} The accuracy of GPS readings is limited, and there can be some minor random shifts.
\end{myenum}

\sstitle{Intuitions} To overcome these challenges, four corresponsive intuitions are employed: 1)~it is hard for us to collect a large scale dataset with illegal parking events, but it is relatively easier to identify negative labels from the dataset (i.e., the normal trajectories) with some heuristics. 2)~comparing to the complicated trajectory characteristics with illegal parking events, the trajectory characteristics without illegal parking events are more stable and easier to identify; 3)~Instead of testing a trajectory individually to see if it is impacted by an illegal parking event, we aggregate all of the trajectories in a time period (i.e., one hour), and extract the overall features; and 4)~as the GPS accuracy is highly related with the build-up proximity~\cite{MKH06}, we build the baseline models to describe the normal trajectory features on each individual road segment.

\sstitle{Main Ideas} With the above intuitions, the shift distribution of the aggregated trajectories is used as the feature to evaluate the existence of illegal parking events on a road segment, as it is a direct result and significantly obvious from our intuition validation experiments in Figure~\ref{fig:intuition}. 

\begin{figure}[t]
  \begin{center}
    \includegraphics[width=3.4in]{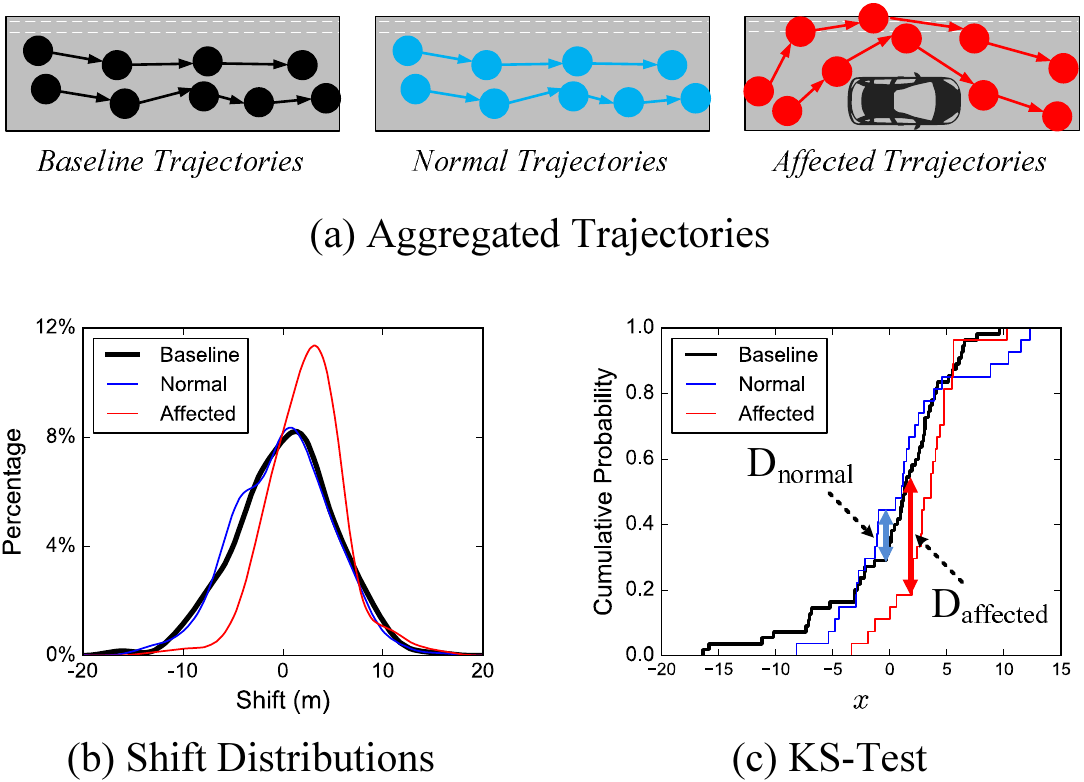}
    \caption{Main Ideas of Illegal Parking Detection.}
  \label{fig:detection}
  \end{center}
   \vspace{-20pt}
\end{figure}

To calculate the difference of shift distributions between the aggregated normal trajectories (or the baseline model) and the evaluation trajectories at the same road segment (i.e., demonstrated in Figure~\ref{fig:detection}a and~\ref{fig:detection}b), the Kolmogorov-Smirnov test (or KS test) is used~\cite{DTZ10}. Figure~\ref{fig:detection}c illustrates the semantic meaning of KS test statistics, where the shift distributions are more similar if there are no illegal parking events on the evaluation trajectories. Then, we set one threshold to determine if the two sets of trajectories are from the same distribution (i.e., evaluation trajectories are in the same scenario as the normal condition) to infer if there are illegal parking events. One threshold is used here, as the impact of trajectory shift from illegal parking events is the same across the whole city (i.e., around the width of a vehicle). Finally, we evaluate the test results of different threshold values to determine the most effective threshold based on the labelled illegal parking events that we collected.

The following sections describe the details on: 1)~building a baseline trajectory distribution model for each road segment; 2)~extracting features from evaluation trajectories; and 3)~performing the distribution test-based evaluation and selecting the threshold to make the detection.

\subsection{Baseline Trajectory Modelling}\label{subsec:modeling}

A baseline trajectory model is built at each road segment to capture the shift distribution for trajectories at the normal scenario (i.e., without illegal parking events). Two heuristics are used:

\sstitle{Naive Baseline Model} A naive baseline uses the shape of a road directly. The assumption of this approach is that if the trajectories travel without the impact of illegal parking events, they should travel perfectly along with the shape of the road segment. 
We use zero-mean Gaussian distribution to simulate the trajectory shifts in normal scenarios.

\sstitle{Night Time Baseline Model} This baseline assumes the bike trajectories at night (e.g., 11:00 PM to 7:00 AM, in our implementation), in most cases, travel without the impact from illegal parking events. To overcome the challenges that 1)~it is possible to have occasional overnight illegal parking events on the street, and 2)~there are very limited number of trajectories traveling during that time period (less than 5\% in the dataset), we aggregated shifts of trajectories on each road segment for a very long time period (i.e., over six months), when constructing baseline models, to minimize the impacts from the above challenges. 

We noted that it is still possible to have ``regular'' illegal parking events during the night time at some road segments, e.g., near a dense residential area. However, we consider them as common knowledge that can be discovered easily and are not the focus of our technique.

\subsection{Evaluation Trajectory Feature Extraction}\label{sec:detection:fea}

This step extracts the features from a set of evaluation trajectories, aggregately, as the individual trajectory is relatively unstable. To ensure a fair sampling from each trajectory in the trajectory set, two tasks are performed: 1)~the trajectories on the road are further segmented (as 50 meters) based on their GPS offsets, to minimize the case that the shift distribution incurred by an illegal parking event is neutralized by the no-illegal parking portion in a very long road segment; 2)~GPS shifts are re-sampled uniformly (e.g., one GPS point every 5 meters) to avoid the case that the shift distribution is dominated by a few highly sampled or very slow trajectories, as there are significant differences between users' behaviors and devices. Then, two methods are proposed to extract the features from evaluation trajectories.

\begin{figure}[h]
  \begin{center}
  \includegraphics[width=3.4in]{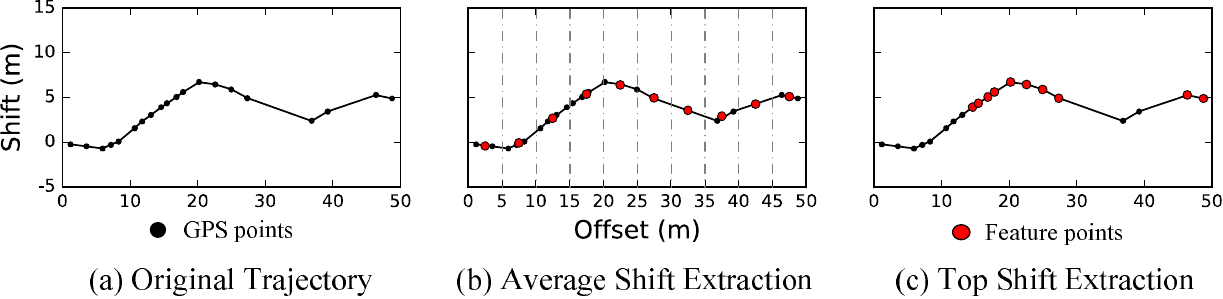}
    \caption{Examples of Trajectory Feature Extraction}
  \label{fig:feature}
  \end{center}
\end{figure}

\sstitle{Average Shift Extraction} This method calculates one average shift value based on all the GPS points within the 5-meters' range of their offsets. As in Figure~\ref{fig:feature}b, the black dots are the original GPS points, and the red dots in each offset range (marked by the dotted vertical lines) are the calculated average shift values. Finally, all calculated shifts are returned as the features.

\sstitle{Top Shift Extraction} This method extracts the top shift values for each trajectory, where the top-10 samples are extracted for each 50-meter road segment in our implementation, as illustrated in Figure~\ref{fig:detection}c. The intuition here is to avoid missing any large shifts caused by a potential illegal parking event. As demonstrated in Figure~\ref{fig:feature}, {\em average shift extraction} doesn't include the highest shift values, which potentially are the results of illegal parking events.

\subsection{Distribution Test-based Detection}\label{sec:detection:kstest}

We use Kolmogorov-Smirnov test (or KS test) statistic on the shift samples from evaluation trajectories and the baseline trajectories to determine if the two samples are drawn from the same distribution. The intuition is that, if the two shift samples are similar enough, then they belong to the same scenario (i.e., without illegal parking events on the road). Otherwise, we consider them as affected by illegal parking events.

\sstitle{KS-Test Statistic Calculation} KS statistic essentially calculates the maximum deviation between two empirical cumulative distribution functions, as demonstrated as the $D_{\text{normal}}$ and $D_{\text{affected}}$ in Figure~\ref{fig:detection}c:
\begin{equation}
  D_{n,m}=\sup_{x}|F_{1,n}(x)-F_{2,m}(x)|
\end{equation}

\noindent where $D_{n,m}$ is the KS statistic, $F_{1,n}$ and $F_{2,m}$ are the empirical cumulative shift distributions of the baseline model and the features of the evaluation trajectories, $m$ and $n$ are the numbers of shift samples, and $sup$ is the supremum function.

\sstitle{Threshold Selection} We reject the assumption that the two samples are from the same distribution (i.e., essentially no illegal parking event) based on the following equation:
\begin{equation}
D_{n,m} > c(\alpha)\sqrt{\frac{n+m}{nm}} \;\& \;  c(\alpha) = \sqrt{-\frac{1}{2}ln(\frac{\alpha}{2})}
\end{equation}

\noindent where $\alpha$ is the probability reflecting if two samples are from the same distribution in KS test. $\alpha$ also is used as the threshold used to reject the assumption that the two samples are from the same distribution (or essentially deciding if the road is with illegal parking events). Instead of using the standard probability threshold, e.g., $\alpha=0.05$, we test a series of different $\alpha$ values based on an illegal parking data set that we collected to select the most effective one. The details of the threshold selection are in Section~\ref{sec:exp:effectiveness}.

%% file: deploy.tex
\section{Deployment of Detection System}\label{sec:deploy}









To improve the response time for detecting the illegal parking events based on massive trajectories over the whole city, the system is deployed based on a parallel computing platform, i.e., Apache Storm. Figure~\ref{fig:clouddeployment} gives an overview of our system deployment on the cloud, with two phases:

\begin{figure}[h]
  \begin{center}
  \includegraphics[width=3.4in]{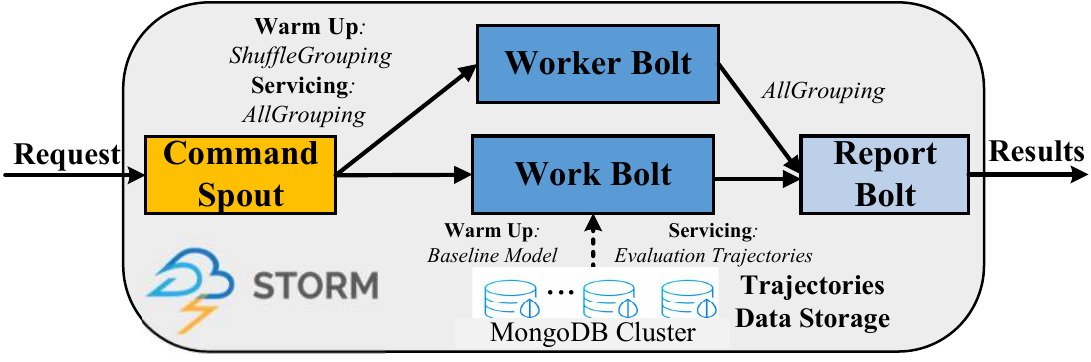}
    \caption{Cloud-based Deployment.}
  \label{fig:clouddeployment}
  \end{center}
\end{figure}

\sstitle{Warm Up} In this phase, the {\em command spout} sends out a set of road segment IDs to the worker nodes using \texttt{ShuffleGrouping} method (i.e., random distribution). For each worker node, they will load the baseline models of the assigned road segments from our trajectory data storage (i.e., MongoDB cluster). After all the baseline models are loaded, the worker node notifies the report bolt. When all the worker bolts are ready, the {\em warm up} phase ends. In this work, the bike trajectories in MongoDB are map-matched offline based on a trajectory preprocessing framework~\cite{RLB18} for mining the historical illegal parking hotspots. Later, by adopting the streaming based map-mathcing system, e.g.,~\cite{BLY+16}, the system can be easily extended to support real-time map-matching and detection in a city.

\sstitle{Servicing} In this phase, a user can input a request to the {\em command spout} to evaluate a set of trajectories in a given temporal range (e.g., the last hour). The temporal range is sent to the {\em worker bolts} with \texttt{AllGrouping} method (i.e., broadcast). For each {\em worker bolt}, it queries the trajectory storage (i.e., MongoDB cluster) to retrieve the trajectories passed during the given temporal range of its preloaded road segment. Then, it performs a KS-test to determine if the road segment has illegal parking event. Finally, the detection results of all the road segments are sent to {\em report bolt} to provide an overall ranking of illegal parking impact at the given time period.

%% file: patrol.tex
\section{Detection-Driven Patrol Scheduling}\label{sec:patrol}
The generated real-time illegal parking detection results in Section~\ref{sec:detection} find the roads with potentially high illegal parking possibilities, which provides us with an opportunity to effectively guide the local polices' enforcement to catch more parking violations under the limited labor effort.
For the government, the schedule decisions can be considered as a sequence of actions. On optimizing the long-term effectiveness of patrol scheduling, to tackle the illegal parking dynamics, we propose to formulate the patrolling task as a Markov Decision Process (MDP), and solves the problem by multi-agent reinforcement learning (MARL).
In the rest of this section,
we first introduce our Markov Decision Process and the MDP formulation of our parking patrol task,
then we elaborate on the patrol scheduling algorithm,
and finally, the design of the patrolling simulator is described.

\subsection{Patrolling Task Formulation}\label{sec:patrol:prelim}

Reinforcement Learning is used to solve the problem of Markov Decision Process, where an agent repeatedly takes action to accumulate reward and transits from the current state to another state randomly. Formally MDP consists of five components $(\mathcal{S}, \mathcal{A},\mathcal{R},\mathcal{P},\gamma)$, representing for the state space, action space, immediate reward function, state transition function and the reward discounting factor. Reinforcement Learning learns policy $\pi$ by optimizing the long-term reward for the given MDP. 

For our parking patrol task, we give the detailed definitions as follows:
\begin{itemize}
\item \textbf{Grid world.} In our problem settings, we partition the urban region into $N_1 \times N_2$ uniform grids, and discretize the time into $10$ minutes' steps. We denote the number of patrol policemen as $N_a$, and the number of time steps as $T$. The positions of patrol policemen and parking violations are also mapped into the partitioned grids. At each time step, the patrol police choose to either ticketing the illegal vehicles in his grid, or move to the surrounding eight grids.
\item \textbf{Agents}. We regard the $N_a$ patrol policemen as homogeneous agents, i.e. the agents follow the same reinforcement learning model. We also assume that the illegal parking information and the location information of all agents are shared.
\item \textbf{State Space $\mathcal{S}$}. For the $k$-th agent, the state is denoted as $s^k_t$ at time step $t$. We consider the following factors that may influence the patrol decision making: 1)~The illegal parking matrix $M$, where each entry $M_{i,j}$ maintains the number of illegal parking events; 2)~The location of the $k$-th agent, which is encoded as one-hot matrix; 3)~The location matrix of partner agents, where each entry indicates the number of other agents in the grid; 4)~The one-hot embedding of the current time step $t$. As a result we have $s^k_t \in \mathbb{R}^{3\times N_1 \times N_2 + T}$.
\item \textbf{Action Space $\mathcal{A}$}. At each time step $t$, the agent $k$ can choose to either stay in the current grid to process the cars or move to the surrounding eight grids, which sum up to $9$ actions in total. The action of agent-$k$ is denoted as one-hot array $a^k_t \in \mathbb{R}^9$ to indicate the choice.
\item \textbf{Immediate Reward $\mathcal{R}$}. We count the number of processed cars at time step $t$ by agent-$k$ as the immediate reward for agent-$k$.
\item \textbf{State Transition $\mathcal{P}$}. As the patrol policemen move or process the illegal parking events according to their actions, the state $s_t$ is update to $s_{t+1}$ via state transition probability function $\mathcal{P}: \mathcal{S} \times \mathcal{A} \times \mathcal{S} \rightarrow \mathbb{R}$. In detail, for each agent $k$, the state transition is conducted by the four steps: 1)~Update illegal parking matrix $M$, as there are illegal parking events emerged, escaped, or processed by the agents in this time step; 2)~Update his own location state; 3)~Update the partner location state according to the new locations of other partners; 4)~Update the time state to $t+1$.
\item \textbf{Policy $\pi$}. The policy $\pi$ is a function that guide the agent to take action based on the state, given the current state, it returns the probability of the next action, which is formally $\pi: \mathcal{S} \times \mathcal{A} \rightarrow \mathbb{R}$.
\end{itemize}

\subsection{Patrol Scheduling Algorithm}

\begin{figure}[b]
\vspace{-15pt}
\begin{center}
\includegraphics[width=3.4in]{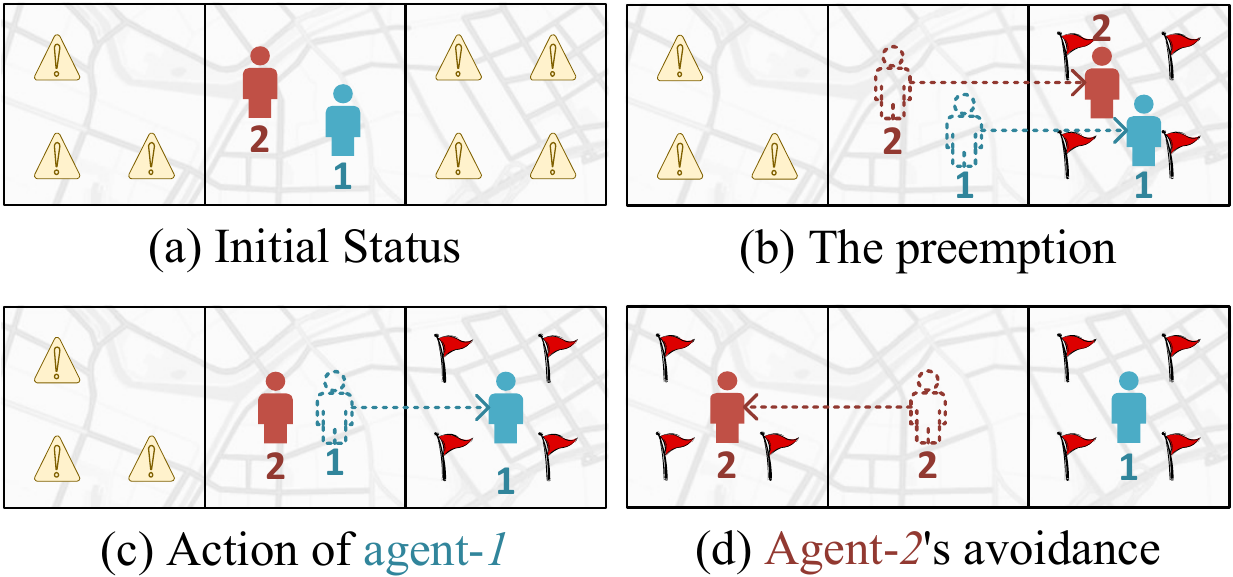}
\caption{Collaboration based on Ordered Action Generation.}
\label{fig:patrol:convention}
\end{center}
\end{figure}

We start by introducing the basic Q-learning. Given an MDP, it learns to evaluate the maximum long-term cumulative discounting reward of state-action combination, i.e. the Q-value function, which is formally:
\begin{equation}
Q_{\pi}(s_t, a_t) = \max_{\pi} \mathbb{E}_{\pi} \left[ \sum_{i=0}^{T-t} \gamma^i \cdot r_{t+i} \right].
\end{equation}
In Q-learning, Bellman Equation (Eq.~\ref{eqn:bellman}) is usually used to estimate the Q-values:
\begin{equation}~\label{eqn:bellman}
Q(s_t, a_t) = r_t + \gamma \cdot \max_{a}{Q(s_{t+1}, a)},
\end{equation}
which essentially follows the best-action policy that maximize Q-value at $t+1$. Since the state space $\mathcal{S}$ and the actions space $\mathcal{A}$ can be very large, DQN~\cite{DQN} uses Deep Neural Network to learn the Q-function $Q(s,a;\theta)$ where $\theta$ is the parameter of the Neural Network. As a result, the Q-function is learned by minimizing the following loss function:
\footnotesize
\begin{equation}~\label{eqn:loss-basic}
J(\theta) = \mathop{\mathbb{E}}_{s_t,a_t, r_t,s_{t+1}} { \left|\left| Q(s_t,a_t;\theta) - (r_t + \gamma \cdot \max_{a}{Q(s_{t+1}, a; \theta')}) \right|\right| _ 2},
\end{equation}
\normalsize
where $(s_t,a_t,r_t,s_{t+1})$ are the experience collected from the environment, and $\theta'$ is the parameter of target Q-function. In practice, the target Q-function parameter $\theta'$ is kept synchronized with $\theta$ periodically. 

As there are multiple patrol cars, it is computationally infeasible to evaluate the joint-action of all patrol cars, which has the action space of $\mathbb{R}^{9 \times N_a}$. To tackle this challenge, we formulate the problem in a multi-agent way: We regard the $N_a$ patrol cars as homogeneous agents that share a common model and act independently, i.e. the agents take actions based on a commonly shared policy, and maximizes the total reward of his own. Instead of generating the actions of agents at the same time, which may cause bad preemption between agents, we set a priority scheme of agents to evaluate and choose the action of each agent one by one.

Figure~\ref{fig:patrol:convention} explains this idea by a toy example, where in Figure~\ref{fig:patrol:convention}a there are two patrol cars in the middle grid and $3$ and $4$ illegal parking events respectively at left and right sides, and we assume the agent can process four events per step. In the case that agents take actions simultaneously, both of them greedily move right since they have symmetric state and share common policy, which causes their preemption on the right grid, and only $4$ events are processed (Figure~\ref{fig:patrol:convention}b). While in contrast in Figure~\ref{fig:patrol:convention}c, given that agent-$1$ takes action ahead of agent-$2$ and informs agent-$2$ of his action (moving right), agent-$2$ can avoid preemption against agent-$1$ and moves to another side, which is more valuable, and as a result, up to $7$ events are processed(Figure~\ref{fig:patrol:convention}d).

\begin{algorithm}[t]
\small
\caption{Algorithm for Generating Joint-actions} ~\label{alg:patrol:replay}
\begin{algorithmic}[1]
\Statex {\bf Input:} The current state $\{s_t^k\}_{k=1}^{N_a}$; The exploration rate $\epsilon_r$
\Statex {\bf Output:} Generated joint-action $a_t$.

\LineComment{{\em Initialization}}
\State Remaining agents $C \leftarrow \{1, \dots, N_a\}$~\label{alg:initialization:start}
\State Temporary states $\hat{s}^k \leftarrow \{s_{t}^k\}_{k=1}^{N_a}$
\State Computed joint-action $a_t \leftarrow \{\phi\}_1^{N_a}$
\LineComment{{\em Action generation}}
\While{$|C|>0$}
    \State $\hat{a}^k \leftarrow \argmax_{a}{Q(\hat{s}^k,a)}, \forall k \in C$
    \LineComment{{\em $Q$-value based agent priority}}
	\State Choose agent $k^* \leftarrow \argmax_{k \in C}{Q(\hat{s}^k, \hat{a}^k)}$
	\LineComment{{\em $\epsilon$-greedy action selection}}
	\If{$\texttt{rand()} < \epsilon_r$}
	    \State Replace $\hat{a}^{k^*}$  by random feasible action of agent-$k^*$ 
	\EndIf
	\State Takes $\hat{a}^{k^*}$ for Agent-$k^*$
	\State Re-evaluate $\{\hat{s}^k\}_{k=1}^{N_a}$
	\State $C \leftarrow C - \{k^*\}$
    \State $a_t[k^*] \leftarrow a^{k^*}$
\EndWhile
\State return $a_t$
\end{algorithmic}
\end{algorithm}

Algorithm~\ref{alg:patrol:replay} gives the details of join-action generation of our MARL model. We iteratively generate actions for each agent. Firstly, we select the remaining agent $k^*$ with maximum $Q$-value(Line $5$-$6$), and execute action for agent-$k^*$ following $\epsilon$-greedy~\cite{DQN} strategy to balance the exploration and exploitation(Line $7$-$9$). After the action, we re-evaluate the agent states according to the new position of agent-$k^*$ and the estimated number of parking violations it can process(Line $10$). Finally we erase $k^*$ from the remaining agent set, and add the agent-action to generation result(Line $11$-$12$). The generated join-actions are executed in environment to collect transition experience of each agent, i.e. $(s_t^k, a_t^k, r_t^k, s_{t+1}^k)$, and DQN is learned by optimizing the following loss function:
\footnotesize
\begin{equation}~\label{eqn:loss-marl}
    J_{ma}(\theta) = \mathop{\mathbb{E}}_{s_t^k,a_t^k, r_t^k, s_{t+1}^k}{ \left|\left| Q(s_t^k,a_t^k;\theta) - (r_t^k + \gamma \cdot \max_{a}{Q(s_{t+1}^k, a; \theta')}) \right|\right| _ 2}.
\end{equation}
\normalsize
Note that actually, the {\em $Q$-value based priority} in Algorithm~\ref{alg:patrol:replay}(Line $5$-$6$) can be replaced by the simple agent-ID priority, which is time-efficient while receiving slightly lower performance in our experiment. 

\subsection{Simulator Design}

The reinforcement learning procedure requires interactions with the environment. In urban scenarios, many research works build their simulators to tackle this problem, and we also follow this idea to simulate our urban illegal parking environment. 

\sstitle{Illegal Parking Emergence} On implementing the simulator, it is non-trivial to estimate the emergence frequency of parking violations over the urban regions: {\em Even for the police officers, due to the low patrol frequency an coverage, it is rather difficult to acquire the city-wide illegal parking data for a long period of time}. Realizing this, to our best effort, we propose to simulate the emergence of illegal parking by our detection results using the massive trajectory data. In detail, we compute the illegal parking results for all road segments at each time step according to Section~\ref{sec:detection:kstest}, and we use $\eta_t^e$ to indicate the status of road segment $e$ at time $t$, i.e.
\begin{equation}~\label{eqn:illpark-indicator}
\eta_t^e = \begin{cases}
      1 & \text{, illegal parking in $e$ at $t$;} \\
      0 & \text{, otherwise}.
    \end{cases}
\end{equation}
For the roads without trajectories passing by, we set this indicator to $0$. With the status of each road segments, the simulator maintains $I_{i,j}$ as the set of roads with parking violations on each grid-$(i,j)$, i.e.:
\begin{equation}~\label{eqn:illpark-grid}
I_{i,j} = \{e | \eta_t^e = 1 \textbf{ and } e \text{ in grid-}(i,j)\}.
\end{equation}
We also compute the illegal parking matrix (in Sec.~\ref{sec:patrol:prelim}) as $M_{i,j} = |I_{i,j}|$.

\sstitle{Simulator in an Episode} We take the potential working hours of a day as an episode, which is set as 6:30 AM - 11:30 PM in our experiment. During each time step of $\Delta t = 10$ minutes, the simulator reacts to the joint-actions by Algorithm~\ref{alg:patrol:simu}, which consists of the following stages:

\begin{itemize}
    \item Creating new illegal parking events, which is computed by the latest illegal parking status(Line 1);
    \item Updating agent locations according to their movement actions, and decrease the remaining time $t_{mv}$(Line 3-4). $t_{mv}$ is determined by the grid size, and we empirically set it as $2$ minutes.
    \item Eliminating the illegal parking records processed by the staying agents, and rewarding the agents. The number of processed cars is drawn from a Poisson distribution with frequency $\lambda_p$(Line 6-8). We set $\lambda_p = 1$ in this work;
    \item Update the time by one step(Line 9).
\end{itemize}

\begin{algorithm}[t]
\small
\caption{Algorithm for Simulation} ~\label{alg:patrol:simu}
\begin{algorithmic}[1]
\Statex {\bf Input:} The joint-action $a_t$; The city-wide illegal parking status $\{ \eta_t^e\}$; Process frequency $\lambda_p$ and moving cost $t_{mv}$.

\State Update $I_{i,j}$ by $\{ \eta_t^e\}$ and Eq.~\ref{eqn:illpark-grid}.
\For{$k = 1 \dots N_a$}
	\If{$a_t^k$ is a movement action}
	    \State Update the location of agent-$k$; $\Delta t \leftarrow \Delta t - t_{mv}$
	\Else
	    \State Get $n_k \sim \texttt{poisson}(\Delta t \cdot \lambda)$; $(i,j) \leftarrow$ agent-$k$'s location
	    \State Reward agent-$k$ by $n_k$
	    \State Pop $n_k$ items of $I_{i,j}$ randomly
	\EndIf
\EndFor
\State Update time $t \leftarrow t+1$ 
\end{algorithmic}
\end{algorithm}

%% file: exp.tex
\section{Experiments}\label{sec:exp}

In this section, we conduct extensive experiments to evaluate the effectiveness and efficiency of our system. We first describe the real dataset used in the paper. Then, we give experiment results to select the proper threshold $\alpha$ used in the KS-test. Then, effectiveness comparisons between different baseline solutions are provided. After that, we test the efficiency performance of our system based on different sizes of Storm clusters. Finally, a set of real case studies are presented to demonstrate the effectiveness of our solution.

\subsection{Datasets}

\sstitle{Road Networks} Road network of Beijing, China is from Open Street Map~\footnote{\url{https://www.openstreetmap.org/}} with $377,559$ nodes and $501,462$ edges.

\sstitle{Mobike Trajectories}  Each Mobike trajectory contains a bike ID, a user ID, a temporal range of the trajectory, a pair of start/end locations, and a sequence of intermediate GPS points. 
  
The dataset used in the paper is the full Mobike trajectory data in the City of Beijing, with the time span of 08/01/2017 - 02/08/2018, the spatial distribution is shown in Figure~\ref{fig:opportunity}a~\footnote{Due to our confidential agreement with Mobike, the trajectory dataset was not made public. Besides, the detailed statistics are kept not disclosed as well}.

\sstitle{Illegal Parking Labels} We collected a set of ground truth labels for illegal parking events in both Chaoyang and Haidian District, in Beijing. Each record contains a road ID, a timestamp, a photo, and a label to indicate the presence of illegal parking events.

The dataset covers 32 roads, and spans over $18$ days (12/26/2017 - 12/30/2017 in Haidian, and 01/12/2018 - 02/09/2018 in Chaoyang). Overall, the $454$ ground truth labels are collected, with 159 records labeled as positive (i.e., with illegal parking events).

\sstitle{Simulator Settings} We build simulator based on the trajectory dataset of Chaoyang, which covers approximately $5km \times 5km$. We set the grid size as $0.5km$, and the duration of grid movement $t_{mv}$ as $2$ minutes. For each episode, we randomly select one day of trajectories from the trajectory dataset to generate the illegal parking events, with around $18000$ events per episode.

\subsection{Detection Performance Evaluation}

In this subsection, we evaluate the effectiveness of the proposed detection model. We first introduce the process to select the most effective threshold for KS-test. Then, we compare our algorithm with three baseline methods and we study the effectiveness and time efficiency of our detection model under different settings. Finally, real world case studies for detection results are described.

\subsubsection{Threshold Selection}\label{sec:exp:effectiveness}

In our implementation, we tried all possible threshold probability $\alpha$ in our KS-test from 0 to 1 with the step size of 0.01. To determine the most effective threshold, we test the detection result with the ground truth labels. A F1 score is calculated to reflect the effectiveness of different threshold values. In detail, we count the numbers of True Positives $N_{TP}$ (i.e., correct identification of positive labels), False Positives $N_{FP}$ (i.e., incorrect identification of positive labels), and False Negatives $N_{FN}$ (i.e., incorrect identification of negative labels) to calculate F1-score: 

    \begin{equation}\label{eqn:f1}
      \begin{split}
        P &= N_{TP}/(N_{TP}+N_{FP}),
        \\
        R &= N_{TP}/(N_{TP}+N_{FN}),
        \\
        F_1 &= 2PR/(P+R).
      \end{split}
    \end{equation} 

\begin{figure}[t]
  \begin{center}
  \includegraphics[width=3.4in]{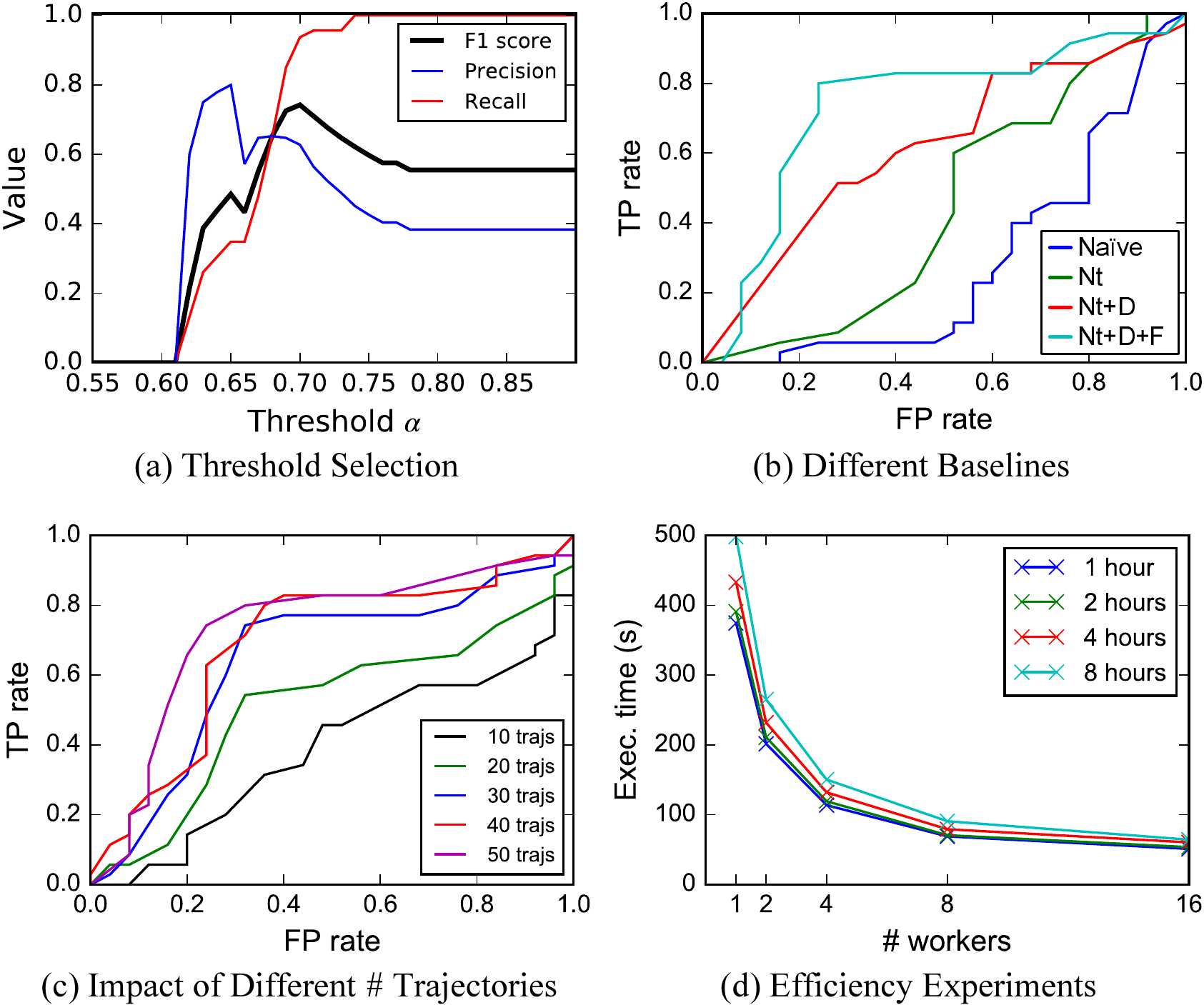}
    \caption{Effectiveness \& Efficiency of Detection.}
  \label{fig:exp:exp}
  \end{center}
  \vspace{-10pt}
\end{figure}

Figure~\ref{fig:exp:exp}a presents the F1 scores with the corresponding \textit{precision} $P$ and \textit{recall} $R$ with respect to different threshold values $\alpha$: 1)~when $\alpha$ is close to zero, all the test data are labeled as negative, so both precision and recall are 0; 2)~with the increase of $\alpha$, more and more instances are labeled as positive, and we identify that $\alpha=0.71$ is the best selection as F1-score reaches the maximum of $0.73$. As a result, we use $\alpha=0.71$ as the threshold in our system.

\subsubsection{Detection Effectiveness Evaluation} 
To achieve a comprehensive comparison with different baseline approaches, we plot the Receiver operating characteristic (ROC curves) for each method. ROC curve illustrates the diagnostic ability of a binary classifier, which plots the true positive rate $TPrate=N_{TP}/(N_{TP}+N_{FN})$ and the false positive rate $FPrate=N_{FP}/(N_{FP}+N_{TN})$ of a model with different parameter settings (i.e., threshold $\alpha$). A method is better, if its AUC (Area Under Curve) is larger.

\sstitle{Baselines} In this experiment, we compare our deployed solution (i.e., \textit{Nt+Dir+T}) with three baseline methods:

{$\bullet$} \textbf{Naive.} It uses the shape of the road segment directly as baseline model, i.e., the shift distribution is set as a Gaussian Distribution with zero mean and $2\sigma$ value is set as $10$ meters. {\em Average shift extraction} is used for processing the evaluation trajectories.

{$\bullet$} \textbf{Nt.} It uses the aggregated night time trajectories to build the baseline model. The features are extracted based on {\em average shift extraction} method.

{$\bullet$} \textbf{Nt+Dir.} In addition to use the night time trajectories as the baseline mode, it also filters the reversed trajectories in each road segment. The trajectory features are extracted based on {\em average shift extraction} method.

{$\bullet$} \textbf{Nt+Dir+T.} It uses the same baseline model as {\em Nt.+Dir}, while {\em top shift extraction} method is employed to extract the features of evaluation trajectories.
 
Figure~\ref{fig:exp:exp}b presents the ROC curves of the 4 baseline methods by varying the threshold $\alpha$ of KS-test. In the measurement of AUC, our deployed method, i.e.,~\textit{Nt+Dir+T} outperforms other three methods significantly, since it considers both directional information and top shift features. The method considering the directional information and removing all the reverse trajectories is much better than the others, which demonstrates the necessity of reverse trajectory removal in the {\em pre-processing} component. Also in the \textit{Naive}, the TP rate is always smaller than FP rate, which implies the inaccuracy of GPS position cannot be neglected.

\sstitle{Different Trajectory Numbers} We study the influence of the detection effectiveness with different numbers of trajectories in the evaluation trajectory feature extraction step. In this experiment, we only choose a portion of labelled road segments with the number of trajectories over 50 in one hour time range, and down samples the dataset randomly to mimic the road with different numbers of trajectories from 10 to 50.

Figure~\ref{fig:exp:exp}c illustrates the ROC curves of our deployed method with different numbers of trajectories used for detection. From the figure we can see that the detection performance increases, when more trajectories can be used. Moreover, when the number of trajectories can be used for detection is less than 20, the detection accuracy is unstable. It confirms the necessity of using the crowd-wisdom to overcome the impacts from skewed individual riding behaviors and the limited GPS accuracy. Finally, the performance of detection model increases slightly when the number of trajectories is over 30. It shows that as long as it has enough number of trajectories (e.g., 30) in an hour on a road segment, our method can provide a relatively accurate detection result.

\subsubsection{Efficiency Evaluation}
The system response time is also tested to show the efficiency of our cloud-based deployment. The experiments are performed in Haidian District, which contains 10\% of the total trajectories in Beijing. We tested the different numbers of requests (i.e., to evaluate 1, 2, 4, and 8 hours of trajectories in the area, the evaluation requests with more than one hour perform the one-hour detection multiple times) with different numbers of worker nodes in Storm.

Figure~\ref{fig:exp:exp}d gives the results of system response time with different settings. We have the following two observations: 1)~the execution time decreases significantly by nearly $50\%$ when adding the number of workers from 1 to 2 and 2 to 4, since the detection tasks are distributed among the worker nodes; and 2)~when the number of workers increases further to 16, the performance gain is much less. This is due to the communication overhead of worker nodes. In terms of different sizes of trajectory data, the response time difference is relatively minor, as the MongoDB cluster provides an efficient trajectory data retrieval interface.

\begin{figure}[h]
  \begin{center}
  \includegraphics[width=3.5in]{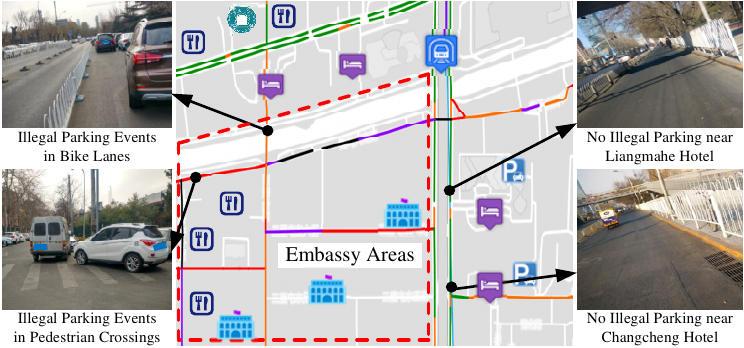}
    \caption{Case Study of Overall Ranking.}
  \label{fig:exp:overallcase}
  \end{center}
  \vspace{-10pt}
\end{figure}

\subsubsection{Case Studies}

We conduct real-world case studies to validate the effectiveness of our detection model. To reflect the severity of the illegal parking events on the roads, we rank them based on the daily average hours with illegal parking events. To validate the correctness of our result, we conduct an on-site case study in the area of Figure~\ref{fig:exp:overallcase}). The area marked with the red dotted lines suffers from severe illegal parking events, based on our calculations. The reason becomes clear, when we got there. It is a very crowded area, with a lot of foreign embassies and high-end restaurants, but with very limited parking spaces. As a result, many people have to leave their cars in the bike lanes or on the pedestrian crossing  (as demonstrated in the left portion of the figure). On the other side, at the east side of the East 3rd ring road, the overall rankings of the green road segments are very low. It is because the POI distribution there is very sparse, with only two large hotels, and the Agriculture Exhibition Center. All of these places are facilitated with very large parking lots. As a result, as shown in the right portion of the figure, the road is clear without any illegal parking events. Note that other events such as road damages can also affect the bike trajectories, which are difficult to distinguish from the true parking violations using the proposed method. The classification between parking cars and other obstacles may be the key challenge in the future. Nevertheless, in this case study, we did not find such events since the events like road damages have a much lower frequency than parking violations.

\subsection{Patrol Effectiveness Evaluation}

\subsubsection{Evaluation Metrics and Baselines}

We choose the straight-forward \underline{R}atio of \underline{P}rocessed Illegal Parking \underline{E}vents metric (RPE) to evaluate the effectiveness of patrolling:
\begin{equation}\label{eqn:rpe}
    \text{RPE} = N_{catch}/ N_{TOT}, 
\end{equation} 
where $N_{catch}$ is the number of processed illegal parking events and $N_{TOT}$ is the total number of events generated by the simulator within an episode. For the evaluation of each method, we repeat the experiment $10$ times and compute the average RPE to overcome the variance of the experiment results. 

In order to validate both the benefit of using detection result as well as the effectiveness of our MARL algorithm, we compare our solution with the following baselines: 
\begin{myitem}
    \item {\bf No-Context Random}. Without the illegal parking detection context, the police car usually patrols randomly in the city. We simulate this scenario by randomly assigning actions to each agent.
    \item {\bf No-Context Ours}, which also excludes the detection context, the algorithm is the same as {\em Ours-aID} described below. It simulates the scenario that the policemen cooperate with each other without our detection information. 
    \item {\bf Heuristic-Greedy}, which is the most na\"ive patrol strategy that leverages our detection result: the agent chooses the neighbor grids with the maximum number of parking violations. This method also simulates an ideal case when the police are experienced and can always go to the area with highest violation risk instead of patrolling randomly. We enhance the patrol with exploration ability by using $\epsilon$-greedy strategy.
    \item {\bf Heuristic-Softmax}, which is similar to {\em Greedy}, but chooses the neighbor grids by the Softmax probability of illegal parking event numbers.
    \item {\bf Independent DQN}, which does not consider the collaboration of agents. Therefore, the location information of other agents are removed from the state space, and the agents take actions simultaneously.
    \item {\bf Ours-aID}, which follows ours Algorithm~\ref{alg:patrol:replay}, but generates actions by agent ID, i.e. from agent-$1$ to agent-$N_a$.
\end{myitem}
During the training stage, all above Deep Q-Networks are built with three dense layers with ReLUs~\cite{relu} as the activation units, where the neuron sizes are set to $256$, $128$, and $32$, respectively. The networks are trained by Adam Optimizer\cite{adam} with the learning rate of $0.001$ and the batch size of $1024$, and the target Q-Network parameters are kept synchronized with the training Q-Network by $2048$ steps periodically. 

\begin{table}[t]
\caption{Effectiveness Evaluation of Patrol Algorithms}
\label{tab:patrol-result}
\vspace{-10pt}
\begin{center}
\begin{tabular}{*4c}
\toprule
\multirow{2}{*}{Methods}   & \multicolumn{3}{c}{Ratio of Processed Events (\textbf{RPE})}  \\ 
                            &     $N_a=6$              &     $N_a=12$              &     $N_a=18$               \\
\midrule
M1. No-Context Random       &     $34.47\%$             &     $63.68\%$             &     $77.16\%$             \\
M2. No-Context Ours         &     $37.84\%$             &     $66.92\%$             &     $79.84\%$             \\
\midrule
M3. Heuristic-Greedy       &     $42.11\%$             &     $64.54\%$             &     $72.90\%$             \\
M4. Heuristic-Softmax       &     $43.96\%$             &     $71.24\%$             &     $85.84\%$             \\
\midrule
M5. Independent DQN         &     $45.18\%$             &     $74.98\%$             &     $87.66\%$             \\
M6. Ours-aID                &     $47.14\%$             &     $77.29\%$             &     $\mathbf{88.06\%}$    \\
\midrule
M7. Ours                    &     $\mathbf{47.82\%}$    &     $\mathbf{78.64\%}$    &     $87.94\%$             \\
\bottomrule
\end{tabular}
\end{center}
\vspace{-10pt}
\end{table}

\subsubsection{Evaluation Results} Table~\ref{tab:patrol-result} gives the evaluation results of the five baselines and our solution under different agent numbers, i.e. $N_a = \{6,12,18\}$. According to Table~\ref{tab:patrol-result}, we mainly have the following four observations:
\begin{itemize}
    \item Without the detection information, the No-Context methods M1\&2 work poorly, while in contrast, even the simple heuristics M4 can out-perform both of them, which implies the benefits of our detection result. Noted that, the results in Table~\ref{tab:patrol-result} assume that the process frequency($\lambda_p$ in Alg.~\ref{alg:patrol:simu}) is the same for all methods, however, M1\&2 may have lower process frequency in practice, since the police are not informed the detailed parking violation risk of each road, and have to inefficiently patrol over the entire roads within the grid. Therefore, M1\&2 may perform even worse in real scenarios, i.e. have lower RPE than in Table~\ref{tab:patrol-result}.
    \item The last three DQN based algorithms perform better than the heuristic solutions(M3\&4). It is intuitive because the DQN based methods consider the optimization of long-term reward. We also find that with the increase of agent number, the performance of the greedy method M3 increases slowly. This is possibly due to the preemption among these agents, the greedy strategy tends to send agents to the same high-value region.
    \item Independent DQN gets relatively lower performance compared to our collaborative MARL method, which may be caused by the lack of collaboration between agents. In addition, we notice that, by applying the heuristic to assign agent priority based on their expected Q-value (M7), the performance increases slightly compared to M6, which shows the influence of action priority in collaboration.
    \item It is interesting to find that, with enough number of agents, e.g. $N_a = 18$, {\em Random} method can also achieve satisfactory result with $REP=77.16\%$. Actually, performances of other methods get very close to each other as well, because of saturation patrol. However, as is pointed out in Introduction, such high-intense patrolling is practically difficult to be applied widely to most cities due to the limited resource. 
\end{itemize}

%% file: related.tex
\section{Related Work}\label{sec:related}

We summarize the related works in three main areas: 1)~urban computing, 2)~trajectory data mining, 3) urban crowdsourcing, and 4) parking patrol.

\sstitle{Urban Computing} Urban computing~\cite{ZCW+14} aims to address different problems in the city. For example, ~\cite{WCL+17,YWK+18,DLL+17} predict the taxi demand to enable smart scheduling and reduce energy wasting, and \cite{ZYZ+15,WL16,YHZ+16,BLZ+16} try to understand human mobility patterns from check-ins. More recently, spatiotemporal models~\cite{wang2019origin,wang2021gallat} based on Graph Neural Networks are proposed, and GNNs are validated to be effective on capturing the spatial correlations between different city regions, which improve the performances on spatiotemporal prediction and inference tasks. Illegal parking detection is also an important issue in urban computing, where most of the existing methods are based on video surveillance~\cite{LRR+09,TFL+11,JWM12}, with limited coverage in a city. However, the system we proposed utilizes a ubiquitous approach. 

\sstitle{Trajectory Data and Mining} Our technique is highly related to trajectory data mining. Based on the massive trajectory data,~\cite{OSZ+14, BHR17} finds popular paths to help planning facilities along the road network. In addition, ~\cite{li2016mining} finds top-$k$ influential locations that cover as many trajectories. In addition, to improve a person's travel experience, trajectory-based solutions for travel time estimation~\cite{WZX14,LGL+17} and reach-ability query~\cite{li2017reachability,WZB+18} are proposed. The closest projects to us are the trajectory anomaly detection, which aims to find trajectories that are dissimilar to the majority of the others. ~\cite{LHK+07,YGC13} are based on classification model, while others~\cite{L16,GXL+11} compare trajectory similarity with history. However, the existing trajectory anomaly detection methods only focus on the high-level route-based difference, while in our problem settings, the difference between trajectories is more subtle at the same road segment.

\sstitle{Urban Crowdsourcing} Essentially, we take the advantage of the massive Mobike users in a city to perform the detection task, which makes us very related to the crowdsourcing techniques. For example,~\cite{TKY17} quantifies the fragility of cities through detecting the delay in commuting activities using GPS data collected from smartphones. ~\cite{QZ16,RCK+10} infer noise levels for locations by smartphone users. ~\cite{JLB+08,HMM13} identify potholes or classify road quality from vehicle's accelerometer data. Different from the above works, we focus on the problem of illegal parking detection. 

\sstitle{Parking Patrol} Parking Patrol plays an important role in urban management. 
\cite{VAT+17,TF+17} studies illegal parking behavior and the relationship between patrolling and parking violations. There are also works study the influence of enforcement policy on illegal parking, e.g.~\cite{CP+92, NR+17}. In terms of parking patrol scheduling,~\cite{KPJ+00} proposes to equip patrol cars with cameras and identify the plate number by computer vision techniques, which improves the efficiency of pasting fine tickets.~\cite{LZO+17,LXL+15} solve the patrol route scheduling problem for parking lots, while in our work, we focus on the parking violations at curbsides all over the city. In addition, similar to our work, some research works also tries to formulate the patrolling task as a reinforcement learning problem, e.g., ~\cite{SRC+04,HZZ+10,LCL+18}. However, these works assume the illegal parking context is known to the agent. In conclusion, none of the above works studies the parking patrol problem without explicit parking context. 

%% file: conclusion.tex
\section{Conclusion}\label{sec:conclusion}

We propose a novel and ubiquitous parking patrol framework, which is driven by real-time illegal parking detection using massive and fine-grained sharing bikes' trajectories. Based on the unique properties of the bike trajectories, we design a comprehensive {\em pre-processing} component to overcome numerous challenges in data cleaning, map-matching, and indexing. In the {\em illegal parking detection} component, we employ a distribution test-based method to determine if the set of evaluation trajectories have the same aggregated shift distribution with the baseline models (i.e., without illegal parking events). {\em Patrol scheduling} component further optimizes the long-term patrol results via collaborative multi-agent reinforcement learning, which considers both the detection results and the cooperation of patrol police. Extensive experiments are performed on a large scale Mobike data. Based on over 400 ground truth labels, our model achieves an F1 score of 0.73, which outperforms all the other baseline approaches. With the help of detection results, in the patrol simulator, our collaborative MARL model improves the traditional untargeted (i.e. No-Context) patrolling by at least 26\% given a limited number of patrol police with $N_a=6$.

%% file: ack.tex
\section*{Acknowledgement}
We thank Beijing Mobike Technology Co., Ltd. (currently Meituan Bike) for providing the sharing bikes' trajectories. 

This work was supported in part by the National Key Research and Development Program of China under Grant 2020YFB1406902, the Key-Area Research and Development Program of Guangdong Province 2020B0101360001, the Shenzhen Science and Technology Research and Development Fundation No.JCYJ20190806143418198 and the National Natural Science Foundation of China (NSFC) under Grant 61872110. 